# The DLV System for Knowledge Representation and Reasoning

NICOLA LEONE and GERALD PFEIFER and WOLFGANG FABER
and
THOMAS EITER and GEORG GOTTLOB and SIMONA PERRI
and
FRANCESCO SCARCELLO

Disjunctive Logic Programming (DLP) is an advanced formalism for knowledge representation and reasoning, which is very expressive in a precise mathematical sense: it allows to express *every* property of finite structures that is decidable in the complexity class $\Sigma_2^P$ ($NP^{NP}$). Thus, under widely believed assumptions, DLP is strictly more expressive than *normal (disjunction-free)* logic programming, whose expressiveness is limited to properties decidable in NP. Importantly, apart from enlarging the class of applications which can be encoded in the language, disjunction often allows for representing problems of lower complexity in a simpler and more natural fashion.

This paper presents the DLV system, which is widely considered the state-of-the-art implementation of disjunctive logic programming, and addresses several aspects. As for problem solving, we provide a formal definition of its kernel language, function-free disjunctive logic programs (also known as disjunctive datalog), extended by weak constraints, which are a powerful tool to express optimization problems. We then illustrate the usage of DLV as a tool for knowledge representation and reasoning, describing a new declarative programming methodology which allows one to encode complex problems (up to $\Delta_3^P$-complete problems) in a declarative fashion. On the foundational side, we provide a detailed analysis of the computational complexity of the language of DLV, and by deriving new complexity results we chart a complete picture of the complexity of this language and important fragments thereof.

Furthermore, we illustrate the general architecture of the DLV system which has been influenced by these results. As for applications, we overview several application front-ends which have been developed on top of DLV to solve specific knowledge representation tasks, and we briefly describe the main international projects investigating the potential of the system for industrial exploitation. Finally, we report about thorough experimentation and benchmarking, which has been carried out to assess the efficiency of the system. The experimental results confirm the solidity of DLV and highlight its potential for emerging application areas like knowledge management and information integration.

This work was supported by Austrian Science Fund (FWF) project Z29-N04 and the European Commission under projects IST-2002-33570 INFOMIX, IST-2001-32429 ICONS, and IST-2001-37004 WASP.

Authors' present addresses: N. Leone, S. Perri – Department of Mathematics, University of Calabria, 87030 Rende (CS), Italy, email: {leone,perri}@mat.unical.it. G. Pfeifer, W. Faber, T. Eiter, and G. Gottlob – Institut für Informationssysteme, Technische Universität Wien, Favoritenstraße 9-11, A-1040 Wien, Austria, email: {pfeifer,gottlob}@dbai.tuwien.ac.at, {faber,eiter}@kr.tuwien.ac.at. F. Scarcello – DEIS Department, University of Calabria, 87030 Rende (CS), Italy, email: scarcello@deis.unical.it.

Contents

## 1. INTRODUCTION

The need for representing and manipulating complex knowledge arising in Artificial Intelligence and in other emerging areas, like Knowledge Management and Information Integration (see also Section 8), has renewed the interest in advanced logic-based formalisms for Knowledge Representation and Reasoning (KR&R), which started in the early 1980s. Among them, Disjunctive Logic Programming (DLP), which has first been considered by Minker [Minker 1982] in the deductive database context, is one of the most expressive KR&R formalisms.

Disjunctive logic programs are logic programs where disjunction is allowed in the heads of the rules and negation may occur in the bodies of the rules. Such programs are now widely recognized as a valuable tool for knowledge representation and commonsense reasoning [Baral and Gelfond 1994; Lobo et al. 1992; Wolfinger 1994; Eiter et al. 1999; Gelfond and Lifschitz 1991; Lifschitz 1996; Minker 1994; Baral 2002]. One of the attractions of disjunctive logic programming (DLP) is its capability of allowing the natural modeling of incomplete knowledge [Baral and Gelfond 1994; Lobo et al. 1992]. Much research has been spent on the semantics of disjunctive logic programs, and several alternative semantics have been proposed [Brass and Dix 1995; Eiter et al. 1997c; Gelfond and Lifschitz 1991; Minker 1982; Przymusinski 1990; 1991; 1995; Ross 1990; Sakama 1989] (see [Apt and Bol 1994; Dix 1995; Lobo et al. 1992; Minker 1994; 1996] for comprehensive surveys). The most widely accepted semantics is the *answer sets semantics* proposed by Gelfond and Lifschitz [1991] as an extension of the stable model semantics of normal logic programs [Gelfond and Lifschitz 1988]. According to this semantics, a disjunctive logic program may have several alternative models (but possibly none), called *answer sets*, each corresponding to a possible view of the world.

Disjunctive logic programs under answer sets semantics are very expressive. It was shown in [Eiter et al. 1997b; Gottlob 1994] that, under this semantics, disjunctive logic programs capture the complexity class $\Sigma_2^P$ (i.e., they allow us to express, in a precise mathematical sense, *every* property of finite structures over a function-free first-order structure that is decidable in nondeterministic polynomial time with an oracle in NP). As Eiter *et al.* [1997b] showed, the expressiveness of disjunctive logic programming has practical implications, since relevant practical problems can be represented by disjunctive logic programs, while they cannot be expressed by logic programs without disjunctions, given current complexity beliefs. The high expressiveness of disjunctive logic programming comes at the price of a higher computational cost in the worst case: The typical reasoning tasks, brave reasoning and cautious reasoning, on disjunctive logic programs are complete problems for the complexity class $\Sigma_2^P$ and $\Pi_2^P$, respectively [Eiter et al. 1997b; Eiter and Gottlob 1995], while non-disjunctive logic programs are complete for NP and co-NP, respectively (cf. [Marek and Truszczyński 1991; Dantsin et al. 2001]).

The hardness of the evaluation of DLP programs has discouraged the implementation of DLP engines for quite some time. Only in the 1990s, more systematic implementation efforts have been launched, including [Fernández and Minker 1992; Lobo et al. 1992; Seipel and Thöne 1994; Dix and Furbach 1996], but the systems created did not reach a level of maturity such that they could be used beyond toy examples.

A first solid, efficiency-geared implementation of a DLP system, called DLV, became available only in 1997, after 15 years of theoretical research on DLP. The development of DLV started at the end of 1996, in a research project funded by the Austrian Science Funds (FWF) and led by Nicola Leone at the Vienna University of Technology; at present, DLV is the subject of an international cooperation between the University of Calabria and the Vienna University of Technology, and its extension and application is supported also by the European Commission through IST projects (see Section 8).

After its first release, the DLV system has been significantly improved over and over in the last years, and its language has been enriched in several ways (see, e.g., [Buccafurri et al. 2000; Buccafurri et al. 2002]). Relevant optimization techniques have been incorporated in all modules of the DLV engine, including database techniques for efficient instantiation [Faber et al. 1999; Leone et al. 2001], novel techniques for answer-set checking [Koch and Leone 1999], heuristics and advanced pruning operators for model generation [Faber et al. 2001; Calimeri et al. 2002].A s a result, at the time being, DLV is generally recognized to be the state-of-the-art implementation of disjunctive logic programming, it is widely used by researchers all over the world, and it is competitive, also from the viewpoint of efficiency, with the most advanced systems in the area of Answer Set Programming (ASP).[1]

This paper focuses on the DLV system, one of the most successful and widely used DLP engines, and provides an in-depth description of several important aspects of the system; it is the first such article, since previous papers focused only on particular issues of the system. The main contributions of this paper are the following.

(1) We provide a formal definition of the kernel language of DLV: disjunctive datalog (i.e., function-free disjunctive logic programs) under the consistent answer sets semantics, extended by weak constraints. The concept of weak constraints presented here generalizes previous work on weak constraints in [Buccafurri et al. 2000], as it allows us to specify both weights and priorities over weak constraints.

(2) We illustrate the usage of DLV as a tool for knowledge representation and reasoning. We describe a declarative "Guess/Check/Optimize" (**GCO**) programming methodology which allows us to encode complex queries and, more generally, search problems in a simple and highly declarative fashion in the DLV language; even some optimization problems of rather high computational complexity (up to $\Delta_3^P$) can be declaratively encoded by using this methodology. We illustrate the usage of the **GCO** methodology on a number of computationally hard problems from various application domains.

(3) We analyze the computational complexity of the language of DLV. The analysis pays attention to the impact of syntactic restrictions on the complexity of the reasoning tasks. We derive new complexity results, which allow for providing a full picture of the complexity of the relevant fragments of the DLV language, and we identify syntactic subclasses which have lower complexity than the whole class. The complexity analysis is in fact the theoretical basis on which the DLV implementation has been founded, and its exploitation is one of the key factors of DLV's efficiency.

(4) We overview application front-ends which have been developed on top of DLV to solve specific KR tasks, and we briefly describe the main international projects aimed at investigating the potential of DLV for industrial exploitation in the areas of Information Integration and Knowledge Management.

---

[1] The term "Answer Set Programming" was coined by Vladimir Lifschitz in his invited talk at ICLP'99, to denote a declarative programming methodology, similar to SAT-based programming, but using more expressive logic programming languages with disjunction or nonmonotonic negation to encode the problems.

(5) We illustrate the general architecture of DLV, explaining the inspiring principles underlying the DLV engine, and we overview the main techniques employed in DLV's implementation, describing how simpler language-fragments are recognized and evaluated by tailored algorithms, exploiting the information coming from the complexity analysis.

(6) We perform a thorough experimentation activity. The main goal of the experiments is to assess the efficiency of DLV among the DLP systems. To this end, we compare DLV with another robust DLP system, named GnT [Janhunen et al. 2000; Janhunen et al. 2003]. We conduct some experiments also in the more general context of Answer Set Programming, comparing DLV and GnT against non-disjunctive ASP systems (such experiments are performed on a smaller set of benchmark problems, since non-disjunctive systems are more limited in power, and they are not able to solve $\Sigma_2^P$-hard problems). In particular, we compare the performance of DLV with two state-of-the-art ASP systems, namely, Smodels [Simons et al. 2002; Niemelä et al. 2000; Niemelä and Simons 1997] and ASSAT [Lin and Zhao 2002].

The entire experimentation activity is conducted on an ample set of benchmarks, covering different application areas, and comprising problems of very different complexities. We provide an in-depth discussion on the benchmark results, highlighting the major strengths and weaknesses of the compared systems, and identifying the best-suited systems for different application and and problem (complexity) classes.

The results of the experiments provide evidence for the wide range of applicability of DLV, and confirm that it is the best-suited ASP system for applications which require dealing with large amounts of data. The latter result is very important, also in the light of the emerging applications of ASP systems in the area of Information Integration in Databases (see Section 8).

The remainder of this paper is structured as follows. In the next section, we described the core language of DLV. After that, we consider in Section 3 knowledge representation and problem solving in DLV, where we present the **GCO** methodology and illustrate it on many examples. Section 4 is devoted to the complexity analysis of the DLV language. Section 5 surveys the DLV front-ends, while Section 6 gives an overview of the architecture and implementation of the DLV system. In Section 7 we then report about extensive experiments and benchmarking. The final Section 8 contains some conclusions, mentions current DLV applications, and gives an outlook to future work.

## 2. CORE LANGUAGE

In this section, we provide a formal definition of the syntax and the semantics of the kernel language of DLV, which is disjunctive datalog under the answer sets semantics [Gelfond and Lifschitz 1991][2] (which involves two kinds of negation), extended with weak constraints. For further background, see [Lobo et al. 1992; Eiter et al. 1997b; Gelfond and Lifschitz 1991].

### 2.1 Syntax

Following Prolog's convention, strings starting with uppercase letters denote variables, while those starting with lower case letters denote constants. In addition, DLV also sup-ports positive integer constants and arbitrary string constants, which are embedded in double quotes. A *term* is either a variable or a constant.

---

[2]See footnote 4.

An *atom* is an expression $p(t_1, \ldots, t_n)$, where $p$ is a *predicate* of arity $n$ and $t_1, \ldots, t_n$ are terms. A *classical literal* $l$ is either an atom $p$ (in this case, it is *positive*), or a negated atom $\neg p$ (in this case, it is *negative*). A *negation as failure (NAF) literal* $\ell$ is of the form $l$ or not $l$, where $l$ is a classical literal; in the former case $\ell$ is *positive*, and in the latter case *negative*. Unless stated otherwise, by *literal* we mean a classical literal.

Given a classical literal $l$, its *complementary* literal $\neg l$ is defined as $\neg p$ if $l = p$ and $p$ if $l = \neg p$. A set $L$ of literals is said to be *consistent* if, for every literal $l \in L$, its complementary literal is not contained in $L$.

Moreover, DLV provides built-in predicates such as the comparative predicates equality, less-than, and greater-than ($=$, $<$, $>$) and arithmetic predicates like addition or multiplication. For details, we refer to [Faber and Pfeifer 1996].

A *disjunctive rule* (*rule*, for short) $r$ is a formula

$$a_1 \ \mathtt{v} \ \cdots \ \mathtt{v} \ a_n :\text{-}\ b_1, \cdots, b_k, \ \text{not } b_{k+1}, \cdots, \ \text{not } b_m. \tag{1}$$

where $a_1, \cdots, a_n, b_1, \cdots, b_m$ are classical literals and $n \geq 0, m \geq k \geq 0$. The disjunction $a_1 \ \mathtt{v} \ \cdots \ \mathtt{v} \ a_n$ is the *head* of $r$, while the conjunction $b_1, \ldots, b_k$, not $b_{k+1}, \ldots,$ not $b_m$ is the *body* of $r$. A rule without head literals (i.e. $n = 0$) is usually referred to as an *integrity constraint*. A rule having precisely one head literal (i.e. $n = 1$) is called a *normal rule*. If the body is empty (i.e. $k = m = 0$), it is called a fact, and we usually omit the "$:\text{-}$" sign.

The following notation will be useful for further discussion. If $r$ is a rule of form (1), then $H(r) = \{a_1, \ldots, a_n\}$ is the set of the literals in the head and $B(r) = B^+(r) \cup B^-(r)$ is the set of the body literals, where $B^+(r)$ (the *positive body*) is $\{b_1, \ldots, b_k\}$ and $B^-(r)$ (*the negative body*) is $\{b_{k+1}, \ldots, b_m\}$.

A *disjunctive datalog program* (alternatively, *disjunctive logic program*, *disjunctive deductive database*) $\mathcal{P}$ is a finite set of rules. A not-free program $\mathcal{P}$ (i.e., such that $\forall r \in \mathcal{P} : B^-(r) = \emptyset$) is called *positive*,[3] and a v-free program $\mathcal{P}$ (i.e., such that $\forall r \in \mathcal{P} : |H(r)| \leq 1$) is called *datalog program* (or *normal logic program*, *deductive database*).

The language of DLV extends disjunctive datalog by another construct, which is called *weak constraint* [Buccafurri et al. 2000]. We define weak constraints as a variant of integrity constraints. In order to differentiate clearly between them, we use for weak constraints the symbol '$:\sim$' instead of '$:\text{-}$'. Additionally, a weight and a priority level of the weak constraint are specified explicitly.

Formally, a weak constraint $wc$ is an expression of the form

$$:\sim \ b_1, \ldots, b_k, \ \text{not } b_{k+1}, \ldots, \ \text{not } b_m. \ [w : l]$$

where for $m \geq k \geq 0$, $b_1, \ldots, b_m$ are classical literals, while $w$ (the *weight*) and $l$ (the *level*, or *layer*) are positive integer constants or variables. For convenience, $w$ and/or $l$ might be omitted and are set to 1 in this case. The sets $B(wc)$, $B^+(wc)$, and $B^-(wc)$ of a weak constraint $wc$ are defined in the same way as for regular integrity constraints.

A DLV *program* $\mathcal{P}$ (*program*, for short) is a finite set of rules (possibly including integrity constraints) and weak constraints. In other words, a DLV program $\mathcal{P}$ is a disjunctive datalog program possibly containing weak constraints. For a program $\mathcal{P}$, let $WC(\mathcal{P})$ denote the set of weak constraints in $\mathcal{P}$, and let $Rules(\mathcal{P})$ denote the set of rules (including

---

[3]In positive programs negation as failure (not) does not occur, while strong negation ($\neg$) may be present.

integrity constraints) in $\mathcal{P}$.

A rule is *safe* if each variable in that rule also appears in at least one positive literal in the body of that rule which is not a comparative built-in. A program is safe, if each of its rules is safe, and in the following we will only consider safe programs.

As usual, a term (an atom, a rule, a program, etc.) is called *ground*, if no variable appears in it. A ground program is also called a *propositional* program.

### 2.2 Semantics

The semantics of DLV programs extends the (consistent) answer sets semantics of disjunctive datalog programs, originally defined in [Gelfond and Lifschitz 1991], to deal with weak constraints. The semantics provided in this section is a slight generalization of the original semantics proposed for weak constraints in [Buccafurri et al. 2000]. In particular, the original language allowed only "prioritized weak constraints", while the DLV language allows both priority levels (layers) and weights for weak constraints.

*Herbrand Universe.* For any program $\mathcal{P}$, let $U_{\mathcal{P}}$ (the Herbrand Universe) be the set of all constants appearing in $\mathcal{P}$. In case no constant appears in $\mathcal{P}$, an arbitrary constant $\psi$ is added to $U_{\mathcal{P}}$.

*Herbrand Literal Base.* For any program $\mathcal{P}$, let $B_{\mathcal{P}}$ be the set of all ground (classical) literals constructible from the predicate symbols appearing in $\mathcal{P}$ and the constants of $U_{\mathcal{P}}$ (note that, for each atom $p$, $B_{\mathcal{P}}$ contains also the strongly negated literal $\neg p$).

*Ground Instantiation.* For any rule $r$, $Ground(r)$ denotes the set of rules obtained by applying all possible substitutions $\sigma$ from the variables in $r$ to elements of $U_{\mathcal{P}}$. In a similar way, given a weak constraint $w$, $Ground(w)$ denotes the set of weak constraints obtained by applying all possible substitutions $\sigma$ from the variables in $w$ to elements of $U_{\mathcal{P}}$. For any program $\mathcal{P}$, $Ground(\mathcal{P})$ denotes the set $GroundRules(\mathcal{P}) \cup GroundWC(\mathcal{P})$, where $GroundRules(\mathcal{P}) = \bigcup_{r \in Rules(\mathcal{P})} Ground(r)$ and $GroundWC(\mathcal{P}) = \bigcup_{w \in WC(\mathcal{P})} Ground(w)$.

Note that for propositional programs, $\mathcal{P} = Ground(\mathcal{P})$ holds.

*Answer Sets.* For every program $\mathcal{P}$, we define its answer sets using its ground instantiation $Ground(\mathcal{P})$ in three steps: First we define the answer sets of positive disjunctive datalog programs, then we give a reduction of disjunctive datalog programs containing negation as failure to positive ones and use it to define answer sets of arbitrary disjunctive datalog programs, possibly containing negation as failure. Finally, we specify the way how weak constraints affect the semantics, defining the semantics of general DLV programs.

An interpretation $I$ is a set of ground classical literals, i.e. $I \subseteq B_{\mathcal{P}}$ w.r.t. a program $\mathcal{P}$. A consistent interpretation $X \subseteq B_{\mathcal{P}}$ is called *closed under* $\mathcal{P}$ (where $\mathcal{P}$ is a positive disjunctive datalog program), if, for every $r \in Ground(\mathcal{P})$, $H(r) \cap X \neq \emptyset$ whenever $B(r) \subseteq X$. An interpretation $X \subseteq B_{\mathcal{P}}$ is an *answer set* for a positive disjunctive datalog program $\mathcal{P}$, if it is minimal (under set inclusion) among all (consistent) interpretations that are closed under $\mathcal{P}$.[4]

EXAMPLE 2.1. The positive program $\mathcal{P}_1 = \{a \vee \neg b \vee c.\}$ has the answer sets $\{a\}$,

---

[4]Note that we only consider *consistent answer sets*, while in [Gelfond and Lifschitz 1991] also the inconsistent set of all possible literals can be a valid answer set.

$\{\neg b\}$, and $\{c\}$. Its extension $\mathcal{P}_2 = \{a \vee \neg b \vee c. \ ; \ :- a.\}$ has the answer sets $\{\neg b\}$ and $\{c\}$. Finally, the positive program $\mathcal{P}_3 = \{a \vee \neg b \vee c. \ ; \ :- a. \ ; \ \neg b :- c. \ ; \ c :- \neg b.\}$ has the single answer set $\{\neg b, c\}$.

The *reduct* or *Gelfond-Lifschitz transform* of a ground program $\mathcal{P}$ w.r.t. a set $X \subseteq B_{\mathcal{P}}$ is the positive ground program $\mathcal{P}^X$, obtained from $\mathcal{P}$ by

—deleting all rules $r \in \mathcal{P}$ for which $B^-(r) \cap X \neq \emptyset$ holds;

—deleting the negative body from the remaining rules.

An *answer set* of a program $\mathcal{P}$ is a set $X \subseteq B_{\mathcal{P}}$ such that $X$ is an answer set of $Ground(\mathcal{P})^X$.

EXAMPLE 2.2. Given the general program $\mathcal{P}_4 = \{a \vee \neg b :- c. \ ; \ \neg b :- \mathrm{not}\ a, \mathrm{not}\ c. \ ; \ a \vee c :- \mathrm{not}\ \neg b.\}$ and $I = \{\neg b\}$, the reduct $\mathcal{P}_4^I$ is $\{a \vee \neg b :- c. \ ; \ \neg b.\}$. It is easy to see that $I$ is an answer set of $\mathcal{P}_4^I$, and for this reason it is also an answer set of $\mathcal{P}_4$.

Now consider $J = \{a\}$. The reduct $\mathcal{P}_4^J$ is $\{a \vee \neg b :- c. \ ; \ a \vee c.\}$ and it can be easily verified that $J$ is an answer set of $\mathcal{P}_4^J$, so it is also an answer set of $\mathcal{P}_4$.

If, on the other hand, we take $K = \{c\}$, the reduct $\mathcal{P}_4^K$ is equal to $\mathcal{P}_4^J$, but $K$ is not an answer set of $\mathcal{P}_4^K$: for the rule $r : a \vee \neg b :- c$, $B(r) \subseteq K$ holds, but $H(r) \cap K \neq \emptyset$ does not. Indeed, it can be verified that $I$ and $J$ are the only answer sets of $\mathcal{P}_4$.

REMARK 2.3. In some cases, it is possible to emulate disjunction by unstratified normal rules by "shifting" the disjunction to the body [Ben-Eliyahu and Dechter 1994; Dix et al. 1996; Leone et al. 1997], as shown in the following example. Consider $\mathcal{P}_5 = \{a \vee b.\}$ and its "shifted version" $\mathcal{P}_5' = \{a :- \mathrm{not}\ b. \ ; \ b :- \mathrm{not}\ a.\}$. Both programs have the same answer sets, namely $\{a\}$ and $\{b\}$.

However, this is not possible in general. For example, consider $\mathcal{P}_6 = \{a \vee b. \ ; \ a :- b. \ ; \ b :- a.\}$. It has $\{a, b\}$ as its single answer set, while its "shifted version" $\mathcal{P}_6' = \{a :- \mathrm{not}\ b. \ ; \ b :- \mathrm{not}\ a. \ ; \ a :- b. \ ; \ b :- a. \}$ has no answer set at all.

Note that these considerations prove that $\mathcal{P}_5$ and $\mathcal{P}_5'$ are not strongly equivalent [Lifschitz et al. 2001]. However, there is no deep relationship between "shifted" programs and strong equivalence: They can be strongly equivalent (e.g. $\mathcal{P}_5 \cup \{:- a, b.\}$ and $\mathcal{P}_5' \cup \{:- a, b.\}$), equivalent (e.g. $\mathcal{P}_5$ and $\mathcal{P}_5'$), or not equivalent at all (e.g. $\mathcal{P}_6$ and $\mathcal{P}_6'$). □

Given a ground program $\mathcal{P}$ with weak constraints $WC(\mathcal{P})$, we are interested in the answer sets of $Rules(\mathcal{P})$ which minimize the sum of weights of the violated (unsatisfied) weak constraints in the highest priority level,[5] and among them those which minimize the sum of weights of the violated weak constraints in the next lower level, etc. Formally, this is expressed by an objective function $H^{\mathcal{P}}(A)$ for $\mathcal{P}$ and an answer set $A$ as follows, using an auxiliary function $f_{\mathcal{P}}$ which maps leveled weights to weights without levels:

$$f_{\mathcal{P}}(1) \ = 1,$$
$$f_{\mathcal{P}}(n) \ = f_{\mathcal{P}}(n-1) \cdot |WC(\mathcal{P})| \cdot w_{max}^{\mathcal{P}} + 1, \quad n > 1,$$
$$H^{\mathcal{P}}(A) = \sum_{i=1}^{l_{max}^{\mathcal{P}}} (f_{\mathcal{P}}(i) \cdot \sum_{w \in N_i^{\mathcal{P}}(A)} weight(w)),$$

where $w_{max}^{\mathcal{P}}$ and $l_{max}^{\mathcal{P}}$ denote the maximum weight and maximum level over the weak constraints in $\mathcal{P}$, respectively; $N_i^{\mathcal{P}}(A)$ denotes the set of the weak constraints in level $i$ that are violated by $A$, and $weight(w)$ denotes the weight of the weak constraint $w$. Note that $|WC(\mathcal{P})| \cdot w_{max}^{\mathcal{P}} + 1$ is greater than the sum of all weights in the program, and therefore guaranteed to be greater than the sum of weights of any single level.

---

[5]Higher values for weights and priority levels mark weak constraints of higher importance. E.g., the most important constraints are those having the highest weight among those with the highest priority level.

Intuitively, the function $f_{\mathcal{P}}$ handles priority levels. It guarantees that the violation of a single constraint of priority level $i$ is more "expensive" then the violation of *all* weak constraints of the lower levels (i.e., all levels $< i$).

For a DLV program $\mathcal{P}$ (possibly with weak constraints), a set $A$ is an *(optimal) answer set* of $\mathcal{P}$ if and only if (1) $A$ is an answer set of $Rules(\mathcal{P})$ and (2) $H^{\mathcal{P}}(A)$ is minimal over all the answer sets of $Rules(\mathcal{P})$.

EXAMPLE 2.4. Consider the following program $\mathcal{P}_{wc}$, which has three weak constraints:

$a \vee b.$
$b \vee c.$
$d \vee \neg d \;\; :\text{-}\;\; a, c.$
$:\sim b. \;\; [1:2]$
$:\sim a, \neg d. \;\; [4:1]$
$:\sim c, d. \;\; [3:1]$

$Rules(\mathcal{P}_{wc})$ admits three answer sets: $A_1 = \{a, c, d\}$, $A_2 = \{a, c, \neg d\}$, and $A_3 = \{b\}$. We have: $H^{\mathcal{P}_{wc}}(A_1) = 3$, $H^{\mathcal{P}_{wc}}(A_2) = 4$, $H^{\mathcal{P}_{wc}}(A_3) = 13$. Thus, the unique (optimal) answer set is $\{a, c, d\}$ with weight 3 in level 1 and weight 0 in level 2.

## 3. KNOWLEDGE REPRESENTATION IN DLV

A main strength of DLV, compared to other answer set programming systems, is its wide range of applicability. While other systems are somehow specialized on a particular class of problems (e.g., NP-complete problems), DLV is more "general purpose" and is able to deal, with a reasonable degree of efficiency, with different kinds of applications, ranging from more "database oriented" deductive database applications (where larger input data have to be dealt with), to NP search and optimization problems, up to harder problems whose complexity resides at the second layer of the Polynomial Hierarchy (more precisely, in $\Sigma_2^P$ and even in $\Delta_3^P$).

In this section, we illustrate the usage of DLV as a tool for knowledge representation and reasoning. We first deal with a couple of classical deductive database applications. Then, we present a new programming methodology, which allows us to encode also hard queries and, more generally, search problems in a simple and highly declarative fashion; even optimization problems of complexity up to $\Delta_3^P$ can be declaratively encoded using this methodology. Finally, we illustrate this methodology on a number of computationally hard problems.

### 3.1 Deductive Database Applications

First, we will present two problems motivated by classical deductive database applications, namely Reachability and Same Generation. Both can be encoded by using only positive datalog rules, thus just scratching the surface of DLV's expressiveness for knowledge representation.

3.1.1 *Reachability.* Given a finite directed graph $G = (N, A)$, we want to compute all pairs of nodes $(a, b) \in N \times N$ such that $b$ is reachable from $a$ through a nonempty sequence of arcs in $A$. In different terms, the problem amounts to computing the transitive closure of the relation $A$.

In the DLV encoding, we assume that $A$ is represented by the binary relation $arc(X, Y)$, where a fact $arc(a, b)$ means that $G$ contains an arc from $a$ to $b$, i.e., $(a, b) \in A$; the set of nodes $N$ is not explicitly represented, since the nodes appearing in the transitive closure are implicitly given by these facts.

The following program then computes a relation $reachable(X, Y)$ containing all facts $reachable(a, b)$ such that $b$ is reachable from $a$ through the arcs of the input graph $G$:

$$reachable(X, Y) \text{:-} \ arc(X, Y).$$
$$reachable(X, Y) \text{:-} \ arc(X, U), reachable(U, Y).$$

3.1.2 *Same Generation.* Given a parent-child relationship (an acyclic directed graph), we want to find all pairs of persons belonging to the same generation. Two persons are of the same generation, if either (i) they are siblings, or (ii) they are children of two persons of the same generation.

The input is provided by a relation $parent(X, Y)$ where a fact $parent(thomas, moritz)$ states that $thomas$ is a parent of $moritz$.

This problem can be encoded by the following program, which computes a relation $samegeneration(X, Y)$ containing all facts such that $X$ is of the same generation as $Y$:

$$samegeneration(X, Y) \text{:-} \ parent(P, X), parent(P, Y).$$
$$samegeneration(X, Y) \text{:-} \ parent(P1, X), parent(P2, Y),$$
$$samegeneration(P1, P2).$$

Unlike Prolog, the order of the rules and the subgoals (literals) in rules bodies does not matter and has no effect on the semantics. In particular, it does not affect the termination of program evaluation, which is not guaranteed under Prolog semantics if, e.g., the last literal in the second rule is reordered to the first position. Like other ASP systems, DLV thus supports a fully modular, rule-by-rule encoding of problems, where the semantics of each rule is independent of its context.

## 3.2   The GCO Declarative Programming Methodology

The core language of DLV can be used to encode problems in a highly declarative fashion, following a Guess/Check/Optimize (**GCO**) paradigm, which is an extension and refinement of the "Guess&Check" methodology in [Eiter et al. 2000a]. We remark that [Lifschitz 2002] introduced a "Generate/Define/Test" methodology closely related to the latter.

In this section, we will first describe the **GCO** technique and we will then illustrate how to apply it on a number of examples. Many problems, also problems of comparatively high computational complexity ($\Sigma_2^P$-complete and $\Delta_3^P$-complete problems), can be solved in a natural manner by using this declarative programming technique. The power of disjunctive rules allows for expressing problems which are more complex than NP, and the (optional) separation of a fixed, non-ground program from an input database allows to do so in a uniform way over varying instances.

Given a set $\mathcal{F}_I$ of facts that specify an instance $I$ of some problem $\mathbf{P}$, a **GCO** program $\mathcal{P}$ for $\mathbf{P}$ consists of the following three main parts:

*Guessing Part.* The guessing part $\mathcal{G} \subseteq \mathcal{P}$ of the program defines the search space, such that answer sets of $\mathcal{G} \cup \mathcal{F}_I$ represent "solution candidates" for $I$.

*Checking Part.* The (optional) checking part $\mathcal{C} \subseteq \mathcal{P}$ of the program filters the solution candidates in such a way that the answer sets of $\mathcal{G} \cup \mathcal{C} \cup \mathcal{F}_I$ represent the admissible solutions for the problem instance $I$.

*Optimization Part.* The (optional) optimization part $\mathcal{O} \subseteq \mathcal{P}$ of the program allows to express a quantitative cost evaluation of solutions by using weak constraints. It implicitly defines an objective function $f : \mathcal{AS}(\mathcal{G} \cup \mathcal{C} \cup \mathcal{F}_I) \to \mathbb{N}$ mapping the answer sets of $\mathcal{G} \cup \mathcal{C} \cup \mathcal{F}_I$ to natural numbers. The semantics of $\mathcal{G} \cup \mathcal{C} \cup \mathcal{F}_I \cup \mathcal{O}$ optimizes $f$ by filtering those answer sets having the minimum value; this way, the optimal (least cost) solutions are computed.

Without imposing restrictions on which rules $\mathcal{G}$ and $\mathcal{C}$ may contain, in the extremal case we might set $\mathcal{G}$ to the full program and let $\mathcal{C}$ be empty, i.e., checking is completely integrated into the guessing part such that solution candidates are always solutions. Also, in general, the generation of the search space may be guarded by some rules, and such rules might be considered more appropriately placed in the guessing part than in the checking part. We do not pursue this issue further here, and thus also refrain from giving a formal definition of how to separate a program into a guessing and a checking part.

In general, both $\mathcal{G}$ and $\mathcal{C}$ may be arbitrary collections of rules (and, for the optimization part, weak constraints), and it depends on the complexity of the problem at hand which kinds of rules are needed to realize these parts (in particular, the checking part).

3.2.1 *Problems in NP and $\Delta_2^P$.* For problems with complexity in NP or, in case of optimization problems, $\Delta_2^P$, often a natural **GCO** program can be designed with the three parts clearly separated into the following simple layered structure:

—The guessing part $\mathcal{G}$ consists of disjunctive rules that "guess" a solution candidate $S$.

—The checking part $\mathcal{C}$ consists of integrity constraints that check the admissibility of $S$.

—The optimization part $\mathcal{O}$ consists of weak constraints.

Each layer may have further auxiliary predicates, defined by normal stratified rules (see Section 4.2 for a definition of stratification), for local computations. This enables, e.g., a more "educated guess" for certain predicates such that unnecessary guesses are eliminated, as we will see in some examples of Subsection 3.3 (HAMPATH, TSP).

The disjunctive rules define the search space in which rule applications are branching points, while the integrity constraints prune illegal branches. Apart from the point in Remark 2.3 regarding the semantic difference between disjunctive rules and equivalent unstratified rules, disjunctive rules are more compact and usually have a more natural reading. The weak constraints in $\mathcal{O}$ induce a modular ordering on the answer sets, allowing the user to specify the best solutions according to an optimization function $f$. With normal stratified rules and DLV's built-in ordering, all polynomial-time functions $f$ can be expressed.

3.2.2 *Problems beyond $\Delta_2^P$.* For problems which are beyond $\Delta_2^P$, and in particular for $\Sigma_2^P$-complete problems, the layered program schema above no longer applies. If $\mathcal{G}$ has complexity in NP, which is the case if disjunction is just used for making the guess and the layer is head-cycle free [Ben-Eliyahu and Dechter 1994], then an answer set $A$ of $\mathcal{G} \cup \mathcal{F}_I$ can be guessed in polynomial time. Hence, it can be shown easily that computing an answer set of the whole program, $\mathcal{G} \cup \mathcal{C} \cup \mathcal{F}_I \cup \mathcal{O}$, is feasible in polynomial time with an NP oracle. Thus, applicability of the same schema to $\Sigma_2^P$-hard problems would imply $\Sigma_2^P \subseteq \Delta_2^P$, which is widely believed to be false.

Until now we tacitly assumed an intuitive layering of the program parts, such that the checking part $\mathcal{C}$ has no "influence" or "feedback" on the guessing part $\mathcal{G}$, in terms of literals which are derived in $\mathcal{C}$ and invalidate the application of rules in $\mathcal{G}$, or make further rules in $\mathcal{G}$ applicable (and thus change the guess). This can be formalized in terms of a "potentially uses" relation [Eiter et al. 1997b] or a "splitting set" condition [Lifschitz and Turner 1994]. Complexity-wise, this can be relaxed to the property that the union of the program parts is head-cycle free.

In summary, if a program solves a $\Sigma_2^P$-complete problem and has guessing and checking parts $\mathcal{G}$ and $\mathcal{C}$, respectively, with complexities below $\Sigma_2^P$, then $\mathcal{C}$ must either contain disjunctive rules or interfere with $\mathcal{G}$ (and in particular head-cycles must be present in $\mathcal{G} \cup \mathcal{C}$).

We close this subsection with remarking that the **GCO** programming methodology has also positive implications from the Software Engineering viewpoint. Indeed, the modular program structure in **GCO** allows for developing programs incrementally, which is helpful to simplify testing and debugging. One can start by writing the guessing part $\mathcal{G}$ and testing that $\mathcal{G} \cup \mathcal{F}_I$ correctly defines the search space. Then, one adds the checking part and verifies that the answer sets of $\mathcal{G} \cup \mathcal{C} \cup \mathcal{F}_I$ encode the admissible solutions. Finally, one tests that $\mathcal{G} \cup \mathcal{C} \cup \mathcal{F}_I \cup \mathcal{O}$ generates the optimal solutions of the problem at hand.

### 3.3 Applications of the GCO Programming Technique

In this section, we illustrate the declarative programming methodology described in Section 3.2 by showing its application on a number of concrete examples.

3.3.1 *Exams Scheduling.* Let us start by a simple scheduling problem. Here we have to schedule the exams for several university courses in three time slots $t_1$, $t_2$, and $t_3$ at the end of the semester. In other words, each course should be assigned exactly one of these three time slots. Specific instances $I$ of this problem are provided by sets $\mathcal{F}_I$ of facts specifying the exams to be scheduled. The predicate $exam$ has four arguments representing, respectively, the *identifier* of the exam, the *professor* who is responsible for the exam, the *curriculum* to which the exam belongs, and the *year* in which the exam has to be given in the curriculum.

Several exams can be assigned to the same time slot (the number of available rooms is sufficiently high), but the scheduling has to respect the following specifications:

$S1$  Two exams given by the same professor cannot run in parallel, i.e., in the same time slot.

$S2$  Exams of the same curriculum should be assigned to different time slots, if possible. If $S2$ is unsatisfiable for all exams of a curriculum $C$, one should:

  $(S2_1)$  first of all, minimize the overlap between exams of the same year of $C$,

  $(S2_2)$  then, minimize the overlap between exams of different years of $C$.

This problem can be encoded in the DLV language by the following **GCO** program $\mathcal{P}_{sch}$:

$$assign(Id, t_1) \lor assign(Id, t_2) \lor assign(Id, t_3) :- exam(Id, P, C, Y). \quad \left.\right\} \text{ Guess}$$

$$:- assign(Id, T), assign(Id', T),$$
$$Id <> Id', exam(Id, P, C, Y), exam(Id', P, C', Y'). \quad \left.\right\} \text{ Check}$$

$$:\sim assign(Id, T), assign(Id', T)$$
$$exam(Id, P, C, Y), exam(Id', P', C, Y), Id <> Id'. \; [: 2]$$
$$:\sim assign(Id, T), assign(Id', T)$$
$$exam(Id, P, C, Y), exam(Id', P', C, Y'), Y <> Y'. \; [: 1]$$

$\left.\begin{array}{c}\\ \\ \\ \\\end{array}\right\}$ **Optimize**

The guessing part $\mathcal{G}$ has a single disjunctive rule defining the search space. It is evident that the answer sets of $\mathcal{G} \cup \mathcal{F}_I$ are the possible assignments of exams to time slots.

The checking part $\mathcal{C}$ consists of one integrity constraint, discarding the assignments of the same time slot to two exams of the same professor. The answer sets of $\mathcal{G} \cup \mathcal{C} \cup \mathcal{F}_I$ correspond precisely to the admissible solutions, that is, to all assignments which satisfy the constraint $S1$.

Finally, the optimizing part $\mathcal{O}$ consists of two weak constraints with different priorities. Both weak constraints state that exams of the same curriculum should *possibly not* be assigned to the same time slot. However, the first one, which has higher priority (level 2), states this desire for the exams of the curriculum of the same year, while the second one, which has lower priority (level 1) states it for the exams of the curriculum of different years. The semantics of weak constraints, as given in Section 2.2, implies that $\mathcal{O}$ captures precisely the constraints $S2$ of the scheduling problem specification. Thus, the answer sets of $\mathcal{G} \cup \mathcal{C} \cup \mathcal{F}_I \cup \mathcal{O}$ correspond precisely to the desired schedules.

3.3.2 *Hamiltonian Path.* Let us now consider a classical NP-complete problem in graph theory, namely Hamiltonian Path.

DEFINITION 3.1 HAMPATH. *Given a directed graph $G = (V, E)$ and a node $a \in V$ of this graph, does there exist a path in $G$ starting at $a$ and passing through each node in $V$ exactly once?*

Suppose that the graph $G$ is specified by using facts over predicates $node$ (unary) and $arc$ (binary), and the starting node $a$ is specified by the predicate $start$ (unary). Then, the following **GCO** program $\mathcal{P}_{hp}$ solves the problem HAMPATH (no optimization part is needed here):

$$inPath(X, Y) \lor outPath(X, Y) :- start(X), arc(X, Y).$$
$$inPath(X, Y) \lor outPath(X, Y) :- reached(X), arc(X, Y).$$
$$reached(X) :- inPath(Y, X). \qquad \textbf{(aux.)}$$

$\left.\begin{array}{c}\\ \\ \\\end{array}\right\}$ **Guess**

$$:- inPath(X, Y), inPath(X, Y1), Y <> Y1.$$
$$:- inPath(X, Y), inPath(X1, Y), X <> X1.$$
$$:- node(X), not\ reached(X), not\ start(X).$$

$\left.\begin{array}{c}\\ \\ \\\end{array}\right\}$ **Check**

The two disjunctive rules guess a subset $S$ of the arcs to be in the path, while the rest of the program checks whether $S$ constitutes a Hamiltonian Path. Here, an auxiliary predicate $reached$ is used, which is associated with the guessed predicate $inPath$ using the last rule. Note that $reached$ is completely determined by the guess for $inPath$, and no further guessing is needed: from the outside, this assignment to $reached$ is an "educated guess" for this predicate, which is more appealing than a blind guess, using a rule "$reached(X) \lor \neg reached(X) :- node(X).$" and subsequently checking whether $reached$ is compatible with the guess for $inPath$.

In turn, through the second rule, the predicate $reached$ influences the guess of $inPath$, which is made somehow inductively: Initially, a guess on an arc leaving the starting node is made by the first rule, followed by repeated guesses of arcs leaving from reached nodes by the second rule, until all reached nodes have been handled.

In the checking part, the first two constraints ensure that the set of arcs $S$ selected by $inPath$ meets the following requirements, which any Hamiltonian Path must satisfy: (i) there must not be two arcs starting at the same node, and (ii) there must not be two arcs ending in the same node. The third constraint enforces that all nodes in the graph are reached from the starting node in the subgraph induced by $S$.

We remark that the above encoding may appear a bit advanced for the unexperienced DLV user. A less sophisticated one guesses for each arc whether it is in the path (i.e., replace the guessing part with a single rule $inPath(X, Y) \vee outPath(X, Y) :- arc(X, Y).$), and defines the predicate $reached$ in the checking part by rules $reached(X) :- start(X).$ and $reached(X) :- reached(Y), inPath(Y, X)..$ However, this encoding is less preferable from a computational point of view, because it leads to a larger search space.

It is easy to see that any set of arcs $S$ which satisfies all three constraints must contain the arcs of a path $v_0, v_1, \ldots, v_k$ in $G$ that starts at node $v_0 = a$, and passes through distinct nodes until no further node is left, or it arrives at the starting node $a$ again. In the latter case, this means that the path is in fact a Hamiltonian Cycle (from which a Hamiltonian path can be immediately computed, by dropping the last arc).

Thus, given a set of facts $\mathcal{F}$ for $node$, $arc$, and $start$, the program $\mathcal{P}_{hp} \cup \mathcal{F}$ has an answer set if and only if the corresponding graph has a Hamiltonian Path. The above program correctly encodes the decision problem of deciding whether a given graph admits a Hamiltonian Path or not.

This encoding is very flexible, and can be easily adapted to solve the *search problems* Hamiltonian Path and Hamiltonian Cycle (where the result has to be a tour, i.e., a closed path). If we want to be sure that the computed result is an *open* path (i.e., it is not a cycle), we can easily impose openness by adding a further constraint $:- start(Y), inPath(\_, Y).$ to the program (like in Prolog, the symbol '$\_$' stands for an anonymous variable whose value is of no interest). Then, the set $S$ of selected arcs in any answer set of $\mathcal{P}_{hp} \cup \mathcal{F}$ constitutes a Hamiltonian Path starting at $a$. If, on the other hand, we want to compute the Hamiltonian cycles, then we just have to strip off the literal $not\ start(X)$ from the last constraint of the program.

3.3.3 *Traveling Salesperson.* The Traveling Salesperson Problem (TSP) is a well-known optimization problem, widely studied in Operation Research.

DEFINITION 3.2 TSP. *Given a weighted directed graph $G = (V, E, C)$ and a node $a \in V$ of this graph, find a minimum-cost cycle (closed path) in $G$ starting at $a$ and passing through each node in $V$ exactly once.*

It is well-known that finding an optimal solution to the Traveling Salesperson Problem (TSP) is intractable. Computing an optimal tour is both NP-hard and co-NP-hard. In fact, in [Papadimitriou 1984] it was shown that deciding whether the cost of an optimal tour is an even number is $\Delta_2^P$-complete.

A DLV encoding for the Traveling Salesperson Problem (TSP) can be easily obtained from an encoding of Hamiltonian Cycle by adding optimization: each arc in the graph carries a weight, and a tour with minimum total weight is selected.

Suppose again that the graph $G$ is specified by predicates $node$ (unary) and $arc$ (ternary), and that the starting node is specified by the predicate $start$ (unary).

We first modify the HAMPATH encoding $\mathcal{P}_{hp}$ in Section 3.3.2 to compute Hamiltonian Cycles, by stripping off literal $\mathrm{not}\ start(X)$ from the last constraint of the program, as explained above. We then add an optimization part consisting of a single weak constraint

$$:\sim inPath(X, Y, C).\ [C:1]$$

which states the preference to avoid taking arcs with high cost in the path, and has the effect of selecting those answer sets for which the total cost of arcs selected by $inPath$ is the minimum.

The full **GCO** program $\mathcal{P}_{tsp}$ solving the TSP problem is thus as follows:

$$
\left.
\begin{aligned}
&inPath(X, Y, C)\ \mathtt{v}\ outPath(X, Y, C) \text{:-}\ start(X), arc(X, Y, C).\\
&inPath(X, Y, C)\ \mathtt{v}\ outPath(X, Y, C) \text{:-}\ reached(X), arc(X, Y, C).\\
&reached(X) \text{:-}\ inPath(Y, X, C).\qquad\qquad\qquad\qquad\textbf{(aux.)}
\end{aligned}
\right\}\ \textbf{Guess}
$$

$$
\left.
\begin{aligned}
&\text{:-}\ inPath(X, Y, \_), inPath(X, Y1, \_), Y <> Y1.\\
&\text{:-}\ inPath(X, Y, \_), inPath(X1, Y, \_), X <> X1.\\
&\text{:-}\ node(X), \mathrm{not}\ reached(X).
\end{aligned}
\right\}\ \textbf{Check}
$$

$$
\left.
\begin{aligned}
&:\sim inPath(X, Y, C).\ [C:1]
\end{aligned}
\right\}\ \textbf{Optimize}
$$

Given a set of facts $\mathcal{F}$ for $node$, $arc$, and $start$ which specifies the input instance, it is easy to see that the (optimal) answer sets of $\mathcal{P}_{tsp} \cup \mathcal{F}$ are in a one-to-one correspondence with the optimal tours of the input graph.

3.3.4   *Ramsey Numbers.*   In the previous examples, we have seen how a search problem can be encoded in a DLV program whose answer sets correspond to the problem solutions. We next see another use of the **GCO** programming technique. We build a DLV program whose answer sets witness that a property does not hold, i.e., the property at hand holds if and only if the DLV program has no answer set. Such a programming scheme is useful to prove the validity of co-NP or $\Pi_2^P$ properties. We next apply the above programming scheme to a well-known problem of number and graph theory.

DEFINITION 3.3 RAMSEY.   *The Ramsey number $R(k, m)$ is the least integer $n$ such that, no matter how we color the arcs of the complete undirected graph (clique) with $n$ nodes using two colors, say red and blue, there is a red clique with $k$ nodes (a red $k$-clique) or a blue clique with $m$ nodes (a blue $m$-clique).*

Ramsey numbers exist for all pairs of positive integers $k$ and $m$ [Radziszowski 1994]. We next show a program $\mathcal{P}_{ramsey}$ that allows us to decide whether a given integer $n$ is <u>not</u> the Ramsey Number $R(3, 4)$. By varying the input number $n$, we can determine $R(3, 4)$, as described below. Let $\mathcal{F}$ be the collection of facts for input predicates *node* and *arc* encoding a complete graph with $n$ nodes. $\mathcal{P}_{ramsey}$ is the following **GCO** program:

$$
\left.
\begin{aligned}
&blue(X, Y)\ \mathtt{v}\ red(X, Y) \text{:-}\ arc(X, Y).
\end{aligned}
\right\}\ \textbf{Guess}
$$

$$
\left.
\begin{aligned}
&\text{:-}\ red(X, Y),\ red(X, Z),\ red(Y, Z).\\
&\text{:-}\ blue(X, Y),\ blue(X, Z), blue(Y, Z),\\
&\quad\ blue(X, W),\ blue(Y, W),\ blue(Z, W).
\end{aligned}
\right\}\ \textbf{Check}
$$

Intuitively, the disjunctive rule guesses a color for each edge. The first constraint eliminates the colorings containing a red clique (i.e., a complete graph) with 3 nodes, and the second constraint eliminates the colorings containing a blue clique with 4 nodes. The program $\mathcal{P}_{ramsey} \cup \mathcal{F}$ has an answer set if and only if there is a coloring of the edges of the complete graph on $n$ nodes containing no red clique of size 3 and no blue clique of size 4. Thus, if there is an answer set for a particular $n$, then $n$ is <u>not</u> $R(3,4)$, that is, $n < R(3,4)$. On the other hand, if $\mathcal{P}_{ramsey} \cup \mathcal{F}$ has no answer set, then $n \geq R(3,4)$. Thus, the smallest $n$ such that no answer set is found is the Ramsey number $R(3,4)$.

The problems considered so far are at the first level of the Polynomial Hierarchy; their complexities, in the ground case, do not exceed $\Delta_2^P$, which is usually placed at the first level, since it is computationally much closer to NP than to $\Sigma_2^P$). We next show that also problems located at the second level of the Polynomial Hierarchy can be encoded by the **GCO** technique.

### 3.3.5 *Quantified Boolean Formulas (2QBF).* 

The first problem at the second level of the Polynomial Hierarchy which we consider is the canonical $\Sigma_2^P$-complete problem 2QBF [Papadimitriou 1994]. Here, we have to decide whether a quantified Boolean formula (QBF) of the shape $\Phi = \exists X \forall Y \phi$, where $X$ and $Y$ are disjoint sets of propositional variables and $\phi = C_1 \vee \ldots \vee C_k$ is a 3DNF formula over $X \cup Y$, evaluates to true. Moreover, in this case, we may want to have a witnessing assignment $\sigma$ to the variables $X$, i.e., an assignment $\sigma$ such that $\phi[X/\sigma(X)]$ is a tautology.[6] This naturally leads to a Guess & Check disjunctive logic program, in which the witness assignment $\sigma$ is guessed by some rules, and the rest of the program is devoted to checking whether $\phi[X/\sigma(X)]$ is a tautology.

Our encoding is a variant of the reduction of 2QBF into a propositional DLP in [Eiter and Gottlob 1995]. Here, a QBF $\Phi$ as above is encoded as a set of facts $F_\Phi$, which is evaluated by a fixed program $\mathcal{P}_{2QBF}$. In detail, $F_\Phi$ contains the following facts:

—$exists(v)$, for each existential variable $v \in X$;

—$forall(v)$, for each universal variable $v \in Y$; and

—$term(p_1, p_2, p_3, q_1, q_2, q_3)$, for each disjunct $l_1 \wedge l_2 \wedge l_3$ in $\phi$, where (i) if $l_i$ is a positive atom $v_i$, then $p_i = v_i$, otherwise $p_i =$ "$true$", and (ii) if $l_i$ is a negated atom $\neg v_i$, then $q_i = v_i$, otherwise $q_i =$"$false$". For example, $term(x_1, true, y_4, false, y_2, false)$, encodes the term $x_1 \wedge \neg y_2 \wedge y_4$.

The program $\mathcal{P}_{2QBF}$ is then

$$
\begin{array}{ll}
\left. \begin{array}{l}
t(true). \quad f(false). \\
t(X) \vee f(X) :\text{-} exists(X).
\end{array} \right\} & \textbf{Guess} \\[2em]
\left. \begin{array}{l}
t(Y) \vee f(Y) \ :\text{-} \ forall(Y). \\
\quad w \ :\text{-} \ term(X, Y, Z, Na, Nb, Nc), \\
\quad\quad\quad\quad t(X), t(Y), t(Z), f(Na), f(Nb), f(Nc). \\
t(Y) \ :\text{-} \ w, forall(Y). \\
f(Y) \ :\text{-} \ w, forall(Y). \\
\quad :\text{-} \ not\ w.
\end{array} \right\} & \textbf{Check}
\end{array}
$$

The guessing part "initializes" the logical constants "$true$" and "$false$" and chooses a witnessing assignment $\sigma$ to the variables in $X$, which leads to an answer set $M_G$ for this part. The more tricky checking part then tests whether $\phi[X/\sigma(X)]$ is a tautology, using a saturation technique [Eiter and Gottlob 1995]: The constraint :- not $w$. enforces that $w$ must be true in any answer set of the program; the preceding two rules imply that such an answer set $M$ contains both $t(y)$ and $f(y)$ for every $y \in Y$. Hence, $M$ has a unique extension with respect to $w$ and all $t(y)$ and $f(y)$ where $y \in Y$. By the minimality of answer sets, an extension of $M_G$ to the (uniquely determined) answer set $M$ of the whole program exists, if and only if for each possible assignment $\mu$ to the variables in $Y$, effected by the disjunctive rule in the checking part, the atom $w$ is derived. The latter holds iff there is some disjunct in $\phi[X/\sigma(X), Y/\mu(Y)]$ which is true. Hence, $M$ is an answer set iff the formula $\phi[X/\sigma(X)]$ is a tautology. In summary, we obtain that $\Phi$ is a Yes-instance, i.e., it evaluates to true, if and only if $\mathcal{P}_{2QBF} \cup F_\Phi$ has some answer set. Moreover, the answer sets of $\mathcal{P}_{2QBF} \cup F_\Phi$ are in one-to-one correspondence with the witnesses $\sigma$ for the truth of $\Phi$.

---

[6]Note that such a witness does no exist for universally quantified formulas of shape $\forall X \exists Y \phi$.

Since 2QBF is $\Sigma_2^P$-complete, as discussed in Section 3.2 the use of disjunction in the checking part is not accidental but necessary: the guessing and checking parts are layered hierarchically (and Splitting Sets [Lifschitz and Turner 1994] do exist).

### 3.3.6   *Strategic Companies.*   A further problem located at the second level of the Polynomial Hierarchy is the following, which is known under the name *Strategic Companies* [Cadoli et al. 1997].

DEFINITION 3.4 STRATCOMP. *Suppose there is a collection $C = \{c_1, \ldots, c_m\}$ of companies $c_i$ owned by a holding, a set $G = \{g_1, \ldots, g_n\}$ of goods, and for each $c_i$ we have a set $G_i \subseteq G$ of goods produced by $c_i$ and a set $O_i \subseteq C$ of companies controlling (owning) $c_i$. $O_i$ is referred to as the* controlling set *of $c_i$. This control can be thought of as a majority in shares; companies not in $C$, which we do not model here, might have shares in companies as well. Note that, in general, a company might have more than one controlling set. Let the holding produce all goods in $G$, i.e. $G = \bigcup_{c_i \in C} G_i$.*

*A subset of the companies $C' \subseteq C$ is a* production-preserving *set if the following conditions hold: (1) The companies in $C'$ produce all goods in $G$, i.e., $\bigcup_{c_i \in C'} G_i = G$. (2) The companies in $C'$ are closed under the controlling relation, i.e. if $O_i \subseteq C'$ for some $i = 1, \ldots, m$ then $c_i \in C'$ must hold.*

*A subset-minimal set $C'$, which is* production-preserving, *is called a* strategic set. *A company $c_i \in C$ is called* strategic, *if it belongs to some strategic set of $C$.*

This notion is relevant when companies should be sold. Indeed, intuitively, selling any non-strategic company does not reduce the economic power of the holding. Computing strategic companies is $\Sigma_2^P$-hard in general [Cadoli et al. 1997]; reformulated as a decision problem ("Given a particular company $c$ in the input, is $c$ strategic?"), it is $\Sigma_2^P$-complete. To our knowledge, it is one of the rare KR problems from the business domain of this complexity that have been considered so far.

In the following, we adopt the setting from [Cadoli et al. 1997] where each product is produced by at most two companies (for each $g \in G$ $|\{c_i \mid g \in G_i\}| \leq 2$) and each company is jointly controlled by at most three other companies, i.e. $|O_i| \leq 3$ for $i = 1, \ldots, m$ (in this case, the problem is still $\Sigma_2^P$-hard). Assume that for a given instance of STRATCOMP, $\mathcal{F}$ contains the following facts:

—$company(c)$ for each $c \in C$,

—$prod\_by(g, c_j, c_k)$, if $\{c_i \mid g \in G_i\} = \{c_j, c_k\}$, where $c_j$ and $c_k$ may possibly coincide,

—$contr\_by(c_i, c_k, c_m, c_n)$, if $c_i \in C$ and $O_i = \{c_k, c_m, c_n\}$, where $c_k$, $c_m$, and $c_n$ are not necessarily distinct.

We next present a program $\mathcal{P}_{strat}$, which solves this hard problem elegantly by only two rules:

$r_{s1} : strat(Y) \lor strat(Z) :\!\text{-} prod\_by(X, Y, Z).$        } **Guess**

$r_{s2} : strat(W) :\!\text{-} contr\_by(W, X, Y, Z), strat(X), strat(Y), strat(Z).$   } **Check**

Here $strat(X)$ means that company $X$ is a strategic company. The guessing part $\mathcal{G}$ of the program consists of the disjunctive rule $r_{s1}$, and the checking part $\mathcal{C}$ consists of the normal rule $r_{s2}$. The program $\mathcal{P}_{strat}$ is surprisingly succinct, given that STRATCOMP is a hard ($\Sigma_2^P$-hard) problem. To overcome the difficulty of the encoding, coming from the intrinsic high complexity of the STRATCOMP problem, we next explain this encoding more in-depth, compared with the previous **GCO** encodings.

The program $\mathcal{P}_{strat}$ exploits the minimization which is inherent to the semantics of answer sets for the check whether a candidate set $C'$ of companies that produces all goods and obeys company control is also minimal with respect to this property.

The guessing rule $r_{s1}$ intuitively selects one of the companies $c_1$ and $c_2$ that produce some item $g$, which is described by $prod\_by(g, c_1, c_2)$. If there were no company control information, minimality of answer sets would naturally ensure that the answer sets of $\mathcal{F} \cup \{r_{s1}\}$ correspond to the strategic sets; no further checking would be needed. However, in case control information is available, the rule $r_{s2}$ checks that no company is sold that would be controlled by other companies in the strategic set, by simply requesting that this company must be strategic as well. The minimality of the strategic sets is automatically ensured by the minimality of answer sets.

The answer sets of $\mathcal{P}_{strat} \cup \mathcal{F}$ correspond one-to-one to the strategic sets of the holding described in $\mathcal{F}$; a company $c$ is thus strategic iff $strat(c)$ is in some answer set of $\mathcal{P}_{strat} \cup \mathcal{F}$.

An important note here is that the checking "constraint" $r_{s2}$ interferes with the guessing rule $r_{s1}$: applying $r_{s2}$ may "spoil" the minimal answer set generated by $r_{s1}$. For example, suppose the guessing part gives rise to a ground rule $r_{sg1}$

$$strat(c1) \lor strat(c2) :\!\text{-} prod\_by(g, c1, c2).$$

and the fact $prod\_by(g, c1, c2)$ is given in $\mathcal{F}$. Now suppose the rule is satisfied in the guessing part by making $strat(c1)$ true. If, however, in the checking part an instance of rule $r_{s2}$ is applied which derives $strat(c2)$, then the application of the rule $r_{sg1}$ to derive $strat(c1)$ is invalidated, as the minimality of answer sets implies that $strat(c1)$ cannot be derived from the rule $r_{sg1}$, if another atom in its head is true.

By the complexity considerations in Subsection 3.2, such interference is needed to solve STRATCOMP in the above way (without disjunctive rules in the Check part), since deciding whether a particular company is strategic is $\Sigma_2^P$-complete. If $\mathcal{P}_{strat}$ is rewritten to eliminate such interference and layer the parts hierarchically, then further disjunctive rules must be added. An encoding which expresses the strategic sets in the generic **GCO**-paradigm with clearly separated guessing and checking parts is given in [Eiter et al. 2000a].

Note that, as described in Remark 2.3, the program above cannot be replaced by a simple normal (non-disjunctive) program. Intuitively, this is due to the fact that disjunction in the head of rules is not exclusive, while at the same time answer sets are subset-minimal. Using techniques like the ones in [Eiter et al. 2003], $\mathcal{P}_{strat}$ can be extended to support an arbitrary number of producers per product and controlling companies per company, respectively.

3.3.7 *Preferred Strategic Companies.* Let us consider an extension of Strategic Companies which also deals with preferences. Suppose that the president of the holding desires, in case of options given by multiple strategic sets, to discard those where certain companies are sold or kept, respectively, by expressing preferences among possible solutions. For example, the president might give highest preference to discard solutions where company $a$ is sold; next important to him is to avoid selling company $b$ while keeping $c$, and of equal importance to avoid selling company $d$, and so on.

In presence of such preferences, the STRATCOMP problem becomes slightly harder, as its complexity increases from $\Sigma_2^P$ to $\Delta_3^P$. Nevertheless, DLV still can handle this quite naturally. Let us assume for simplicity that the president's preferences are represented by a single predicate $avoid(c_{sell}, c_{keep}, pr)$, which intuitively states that selling $c_{sell}$ while keeping $c_{keep}$ should be avoided with priority $pr$; in the above example, the preferences would be $avoid(a, c_\top, top)$, $avoid(b, c, top-1)$, $avoid(d, c_\top, top-1)$, …, where $c_\top$ is a dummy company which belongs to every strategic set, and $top$ is the highest priority number. Then, we can easily represent this more complicated problem, by adding the following weak constraint to the original encoding for STRATCOMP:

$$:\sim avoid(Sell, Keep, Priority), \text{not } strat(Sell), strat(Keep). [: Priority]$$

The (optimal) answer sets of the resulting program then correspond to the solutions of the above problem.

## 4. THE COMPLEXITY OF THE DLV LANGUAGE

In this section, we analyze the computational complexity of the DLV language and some relevant fragments thereof. The exploitation of the analysis of the computational complexity of DLV programs is one of the key factors of DLV's efficiency. Indeed, as we will point out in Section 6, the DLV system recognizes syntactic subclasses of the language and employs "ad hoc" evaluation algorithms if the subclass has a lower complexity.

In the sequel of this section, we first provide some preliminaries on complexity theory. Subsequently, we define a couple of relevant syntactic properties of DLV programs, which allow us to single out computationally simpler subclasses of our language.Then, we define the main computational problems under consideration and derive their precise complexity. We conclude with a discussion, paying attention to the impact of syntactic restrictions.

### 4.1 A Reminder of the Polynomial Hierarchy

We assume that the reader has some acquaintance with the concepts of NP-completeness and complexity theory and provide only a very short reminder of the complexity classes of the Polynomial Hierarchy which are relevant to this section. The book [Papadimitriou 1994] is an excellent source for deepening the knowledge in this field.

The classes $\Sigma_k^P$, $\Pi_k^P$, and $\Delta_k^P$ of the *Polynomial Hierarchy* (PH, cf. [Johnson 1990]) are defined as follows:

$$\Delta_0^P = \Sigma_0^P = \Pi_0^P = P; \text{ and for all } k \geq 1, \Delta_k^P = P^{\Sigma_{k-1}^P}, \Sigma_k^P = NP^{\Sigma_{k-1}^P}, \Pi_k^P = co\text{-}\Sigma_k^P,$$

where $NP^C$ denotes the class of decision problems that are solvable in polynomial time on a nondeterministic Turing machine with an oracle for any decision problem $\pi$ in the class $C$. In particular, $NP = \Sigma_1^P$, co-NP $= \Pi_1^P$, and $\Delta_2^P = P^{NP}$.

The oracle replies to a query in unit time, and thus, roughly speaking, models a call to a subroutine for $\pi$ that is evaluated in unit time.

Observe that for all $k \geq 1$,

$$\Sigma_k^P \subseteq \Delta_{k+1}^P \subseteq \Sigma_{k+1}^P \subseteq \text{PSPACE}$$

where each inclusion is widely conjectured to be strict. By the rightmost inclusion above, all these classes contain only problems that are solvable in polynomial space. They allow, however, a finer grained distinction among NP-hard problems that are in PSPACE.

## 4.2 Relevant Fragments of the DLV Language

In this section, we introduce syntactic classes of DLV programs with a number of useful and interesting properties. First we need the following:

DEFINITION 4.1. Functions $|| \ || : B_{\mathcal{P}} \rightarrow \{0, 1, \ldots\}$ from the ground (classical) literals of the Herbrand Literal Base $B_{\mathcal{P}}$ to finite ordinals are called *level mappings* of $\mathcal{P}$.

Level mappings give us a useful technique for describing various classes of programs.

DEFINITION 4.2. A disjunctive logic program $\mathcal{P}$ is called *(locally) stratified* [Apt et al. 1988; Przymusinski 1988], if there is a level mapping $|| \ ||_s$ of $\mathcal{P}$ such that, for every rule $r$ of $Ground(\mathcal{P})$,

(1) For any $l \in B^+(r)$, and for any $l' \in H(r)$, $||l||_s \leq ||l'||_s$;

(2) For any $l \in B^-(r)$, and for any $l' \in H(r)$, $||l||_s < ||l'||_s$.

(3) For any $l, l' \in H(r)$, $||l||_s = ||l'||_s$.

EXAMPLE 4.3. Consider the following two programs.

$$\mathcal{P}_7 : \ p(a) \vee p(c) :\text{- not } q(a). \qquad \mathcal{P}_8 : \ p(a) \vee p(c) :\text{- not } q(b).$$
$$p(b) :\text{- not } q(b). \qquad\qquad\quad q(b) :\text{- not } p(a).$$

It is easy to see that program $\mathcal{P}_7$ is stratified, while program $\mathcal{P}_8$ is not. A suitable level mapping for $\mathcal{P}_7$ is the following:

$$||p(a)||_s = 2 \quad ||p(b)||_s = 2 \quad ||p(c)||_s = 2 \quad ||q(a)||_s = 1 \quad ||q(b)||_s = 1 \quad ||q(c)||_s = 1$$

As for $\mathcal{P}_8$, an admissible level mapping would need to satisfy $||p(a)||_s < ||q(b)||_s$ and $||q(b)||_s < ||p(a)||_s$, which is impossible.

Another interesting class of problems consists of head-cycle free programs.

DEFINITION 4.4. A program $\mathcal{P}$ is called *head-cycle free (HCF)* [Ben-Eliyahu and Dechter 1994], if there is a level mapping $|| \ ||_h$ of $\mathcal{P}$ such that, for every rule $r$ of $Ground(\mathcal{P})$,

(1) For any $l \in B^+(r)$, and for any $l' \in H(r)$, $||l||_h \leq ||l'||_h$;

(2) For any pair $l, l' \in H(r)$ $||l||_h \neq ||l'||_h$.

EXAMPLE 4.5. Consider the following program $\mathcal{P}_9$.

$$\mathcal{P}_9 : \ a \vee b.$$
$$a :\text{- } b.$$

It is easy to see that $\mathcal{P}_9$ is head-cycle free; an admissible level mapping for $\mathcal{P}_9$ is given by $||a||_h = 2$ and $||b||_h = 1$. Consider now the program

$$\mathcal{P}_{10} = \mathcal{P}_9 \cup \{b :\text{- } a.\}$$

$\mathcal{P}_{10}$ is not head-cycle free, since $a$ and $b$ should belong to the same level by Condition (1) of Definition 4.4, while they cannot by Condition (2) of that definition. Note, however, that $\mathcal{P}_{10}$ is stratified.

### 4.3 Main Problems Considered

As for the classical nonmonotonic formalisms, three important decision problems, corresponding to three different reasoning tasks, arise in the context of the DLV language:

**Brave Reasoning**. Given a program $\mathcal{P}$, and a ground atom $A$, decide whether $A$ is true in some answer set of $\mathcal{P}$ (denoted $\mathcal{P} \models_b A$).

**Cautious Reasoning**. Given a program $\mathcal{P}$, and a ground atom $A$, decide whether $A$ is true in all answer sets of $\mathcal{P}$ (denoted $\mathcal{P} \models_c A$).

**Answer Set Checking.** Given a program $\mathcal{P}$, and a set $M$ of ground literals as input, decide whether $M$ is an answer set of $\mathcal{P}$.

We study the complexity of these decision problems, which are strongly relevant to the tasks performed by the DLV computational engine. Brave Reasoning is strictly related also to the problem of *finding* an answer set, which is to be solved, for instance, when a search problem (like Hamiltonian Path in Section 3) is encoded in a DLV program.

In the following, we analyze the computational complexity of the two decision problems specified above for ground (i.e., propositional) DLV programs; we shall address the case of non-ground programs at the end of this section.

An interesting issue, from the viewpoint of system implementation, is the impact of syntactic restrictions on the logic program $\mathcal{P}$. Starting from normal positive programs (without negation and disjunction), we consider the effect of allowing the (combined) use of the following constructs:

—stratified (nonmonotonic) negation ($\mathrm{not}_s$),

—arbitrary (nonmonotonic) negation ($\mathrm{not}$),

—head-cycle free disjunction ($\mathrm{v_h}$),

—arbitrary disjunction ($\mathrm{v}$),

—weak constraints ($w$).[7]

Given a set $X$ of the above syntactic elements (with at most one negation and at most one disjunction symbol in $X$), we denote by $\mathrm{DLV}[X]$ the fragment of the DLV language where the elements in $X$ are allowed. For instance, $\mathrm{DLV}[\mathrm{v_h}, \mathrm{not}_s]$ denotes the fragment allowing head-cycle free disjunction and stratified negation, but no weak constraints.

### 4.4 Derivation of Complexity Results

The computational complexity of the above problems for a number of fragments of the DLV language has been previously analyzed, cf. [Eiter et al. 1997b; Gottlob 1994; Buccafurri et al. 2000; Eiter et al. 1998b; Eiter and Gottlob 1995]. To obtain a full picture of the complexity of the fragments of the DLV language, we establish in this subsection the complexity characterization of the fragments which have not been studied yet. The reader who is not interested in the technicalities might jump directly to Section 4.5, which provides a summary and discussion of the results.

---

[7]Following [Buccafurri et al. 2000], possible restrictions on the support of negation affect $Rules(\mathcal{P})$, that is, the rules (including the integrity constraints) of the program, while weak constraints, if allowed, can freely contain both positive and negative literals in any fragment of the DLV language we consider.

Throughout this section, we consider the ground case, i.e., we assume that programs and, unless stated otherwise, also atoms, literals etc. are ground. Furthermore, to simplify matters and stay in line with results from the literature, we shall tacitly restrict the language fragments by disregarding strong negation and integrity constraints in programs. However, this is insignificant inasmuch as the results in presence of these constructs are the same (see, e.g., [Buccafurri et al. 2000]). Some remarks on the complexity and expressiveness of non-ground programs are provided at the end of Section 4.5.

We start by analyzing the complexity of cautious reasoning for DLV[ $\mathtt{v}$ , not, $w$], i.e., the full DLV language. To determine the upper complexity bound, we first prove two lemmas.

LEMMA 4.6. *Given a* DLV[ $\mathtt{v}$ , not, $w$] *program* $\mathcal{P}$ *and an integer* $n \geq 0$ *as input, deciding whether some answer set* $M$ *of* $Rules(\mathcal{P})$ *exists such that* $H^{\mathcal{P}}(M) \leq n$ *is in* $\Sigma_2^P$.

PROOF. We can decide the problem as follows. Guess $M \subseteq B_{\mathcal{P}}$, and check that: (1) $M$ is an answer set of $Rules(\mathcal{P})$, and (2) $H^{\mathcal{P}}(M) \leq n$. Clearly, property (2) can be checked in polynomial time, while (1) can be decided by a single call to an NP oracle, cf. [Marek and Truszczyński 1991; Eiter et al. 1997b]. The problem is therefore in $\Sigma_2^P$. □

LEMMA 4.7. *Given a* DLV[$\mathtt{v}$, not, $w$] *program* $\mathcal{P}$, *a positive integer* $n$, *and an atom* $A$ *as input, deciding whether there exists an answer set* $M$ *of* $Rules(\mathcal{P})$ *such that* $H^{\mathcal{P}}(M) = n$ *and* $A \notin M$ *is in* $\Sigma_2^P$.

PROOF. Similar as above, guess $M \subseteq (B_{\mathcal{P}} - \{A\})$ and check that: (1) $M$ is an answer set of $Rules(\mathcal{P})$, and (2) $H^{\mathcal{P}}(M) = n$. Clearly, property (2) can be checked in polynomial time, while (1) again can be decided by a single call to an NP oracle. The problem is therefore in $\Sigma_2^P$. □

We are now in the position to determine the precise complexity of cautious reasoning over full DLV programs.

THEOREM 4.8. *Given a* DLV[$\mathtt{v}$, not, $w$] *program* $\mathcal{P}$, *and an atom* $A$ *as input, deciding whether* $A$ *is true in all answer sets of* $\mathcal{P}$ *is* $\Delta_3^P$-*complete. Hardness holds even if* $\mathcal{P}$ *is a* DLV[$\mathtt{v}$, $w$] *program.*

PROOF. *Membership.* We prove that the complementary problem is in $\Delta_3^P$ as follows.

Given $\mathcal{P}$, let $u = u(\mathcal{P})$ be the value that the objective function $H^{\mathcal{P}}(M)$ from Section 2.2 takes when an interpretation $M$ would violate all weak constraints in $WC(\mathcal{P})$ (in each layer). More precisely, $u(\mathcal{P}) = \sum_{i=1}^{l_{max}^{\mathcal{P}}} (f_{\mathcal{P}}(i) \cdot \sum_{w \in WC_i} weight(w))$, where $WC_i$ is the set of weak constraints in layer $i$ (see Section 2.2).

Clearly, $u$ is an upper bound on the cost $s^*$ of any optimal answer set of $\mathcal{P}$. By a binary search on $[0..u]$, we can compute $s^*$ using an oracle which decides, given $\mathcal{P}$ and an integer $n \geq 0$, the existence of an answer set $M$ of $Rules(\mathcal{P})$ such that $H^{\mathcal{P}}(M) \leq n$ (on the first call, $n = u/2$ ; then if the oracle answers "yes," $n = u/4$; otherwise, $n$ is set to $u/2 + u/4$, and so on, according to standard binary search).

Observe that $u$ is computable in polynomial time but, because of binary number representation, its *value* might be exponential in the size of the input. However, the number of calls to the oracle is logarithmic in $u$, and thus polynomial in the size of the input. Furthermore, the oracle employed is in $\Sigma_2^P$ by virtue of Lemma 4.6. Thus, computing $s^*$ as above is possible in polynomial time with polynomially many calls to a $\Sigma_2^P$ oracle.

Finally, a further call to a $\Sigma_2^P$ oracle verifies that there is an answer set $M$ of $Rules(\mathcal{P})$ such that $A \notin M$ and $H^{\mathcal{P}}(M) = s^*$ (this is feasible in $\Sigma_2^P$ from Lemma 4.7).

In summary, disproving that $A$ is true in all answer sets of $\mathcal{P}$ is in $\Delta_3^P$. Since co-$\Delta_3^P = \Delta_3^P$, cautious reasoning on full DLV programs is in $\Delta_3^P$ as well.

*Hardness.* We reduce brave reasoning on DLV[v, $w$] programs, which was shown to be $\Delta_3^P$-hard in [Buccafurri et al. 2000], to cautious reasoning on DLV[v, $w$] programs.

Given a DLV[v, $w$] program $\mathcal{P}$ and an atom $A$, we build a DLV[v, $w$] program $\mathcal{P}'$ such that $\mathcal{P} \models_b A$ if and only if $\mathcal{P}' \models_c A$.

To that end, we first consider the program $\mathcal{P}_1$, which we obtain from $\mathcal{P}$ by the following transformations: (i) shift up the priority level of each weak constraint by 1; (ii) add the weak constraint :∼ not $A$. [ : 1] (which is then the only weak constraint in the lowest layer). If $\mathcal{P} \not\models_b A$, then nothing changes, and the (optimal) answer sets of $\mathcal{P}_1$ are the (optimal) answer sets of $\mathcal{P}$. If $\mathcal{P} \models_b A$, then the newly added weak constraint filters out those answer sets of $\mathcal{P}$ where $A$ is false. In this case, the optimal answer sets of $\mathcal{P}_1$ are precisely the optimal answer sets of $\mathcal{P}$ containing $A$.

However, the equivalence "$\mathcal{P} \models_b A$ iff $\mathcal{P}_1 \models_c A$" does not hold if $\mathcal{P}$ has no answer set, as $\mathcal{P}_1 \models_c A$ while $\mathcal{P} \not\models_b A$ in this case. To account for this case, we transform $\mathcal{P}_1$ into $\mathcal{P}'$ as follows: (i) add the disjunctive fact $w$ v $w'$., where $w$ and $w'$ are fresh atoms; (ii) add the atom $w'$ to the body of every rule of $\mathcal{P}_1$ (including constraints); and (iii) add the weak constraint :∼ $w$. [ : $top$], where $top$ is higher than the maximum layer of $\mathcal{P}_1$. If $\mathcal{P}$ has some answer set, then the (optimal) answer sets of $\mathcal{P}'$ are precisely the same as the optimal answer sets of $\mathcal{P}_1$ (modulo the atom $w'$, which occurs in each answer set of $\mathcal{P}'$). Otherwise, $\mathcal{P}'$ has precisely one answer set, namely $\{w\}$. We have therefore reached our goal, as $\mathcal{P} \models_b A$ iff $\mathcal{P}' \models_c A$ and $\mathcal{P}'$ is obviously constructible from $\mathcal{P}$ in polynomial time (in fact, in logarithmic space).  □

We next investigate the impact of disallowing positive recursion through disjunction (HCF programs). We first derive the analogs of Lemmas 4.6 and 4.7 for DLV[$v_h$, not, $w$], which follows by a similar proof using the fact that answer set checking for a head-cycle free program can be done in polynomial time [Ben-Eliyahu and Dechter 1994], rather than being co-NP-complete as in the general case.

LEMMA 4.9. *Given a* DLV[$v_h$, not, $w$] *program $\mathcal{P}$, and a positive integer $n$, deciding whether there exists an answer set $M$ of $Rules(\mathcal{P})$ such that $H^{\mathcal{P}}(M) \leq n$ is in NP.*

PROOF. We guess $M \subseteq B_{\mathcal{P}}$, and check that: (1) $M$ is an answer set of $Rules(\mathcal{P})$, and (2) $H^{\mathcal{P}}(M) \leq n$. Both of the above properties can be checked in polynomial time. Indeed, since the program is head-cycle free, answer-set checking is feasible in polynomial time [Ben-Eliyahu and Dechter 1994]. The problem is therefore in NP.  □

LEMMA 4.10. *Given a* DLV[$v_h$, not, $w$] *program $\mathcal{P}$, a positive integer $n$, and an atom $A$ as input, deciding whether there exists an answer set $M$ of $Rules(\mathcal{P})$ such that $A \notin M$ and $H^{\mathcal{P}}(M) = n$ is in NP.*

PROOF. We guess $M \subseteq (B_{\mathcal{P}} - \{A\})$, and check that: (1) $M$ is an answer set of $Rules(\mathcal{P})$, and (2) $H^{\mathcal{P}}(M) = n$. Clearly, properties (1) and (2) can be checked in polynomial time. In particular, for property (1) this follows from the fact that the program is head-cycle free [Ben-Eliyahu and Dechter 1994]. The problem is therefore in NP.  □

THEOREM 4.11. *Given a* DLV[$v_h$, not, $w$] *program* $\mathcal{P}$, *and an atom $A$ as input, deciding whether $A$ is true in all answer sets of* $\mathcal{P}$ *is* $\Delta_2^P$-*complete. Hardness holds even if* $\mathcal{P}$ *is either a* DLV[$v_h$, $w$] *or a* DLV[not, $w$] *program.*

PROOF. *Membership.* To prove that the complementary problem is in $\Delta_2^P$, we proceed as in the membership proof of Theorem 4.8. From Lemma 4.9 and Lemma 4.10, this time the oracle needed is in NP (instead of $\Sigma_2^P$). Therefore, cautious reasoning on DLV[$v_h$, not, $w$] programs is in $\Delta_2^P$.

*Hardness.* We reduce brave reasoning on DLV[$v_h$, $w$] programs, which was shown to be $\Delta_2^P$-hard in [Buccafurri et al. 2000], to cautious reasoning on DLV[$v_h$, $w$] programs. The reduction is precisely the same as in the Hardness proof of Theorem 4.8. Note that the program $\mathcal{P}'$ resulting from the reduction is in DLV[$v_h$, $w$] if the original program $\mathcal{P}$ is in DLV[$v_h$, $w$]: the addition of the disjunctive fact does not affect head-cycle freeness, and negation in weak constraints is allowed in the fragment DLV[$v_h$, $w$].

Concerning the case of DLV[not, $w$], recall that brave reasoning on DLV[not, $w$] is also $\Delta_2^P$-hard [Buccafurri et al. 2000]. Now, assume that $\mathcal{P}$ is a DLV[not, $w$] program, and apply again the same reduction as in the hardness part of the proof of Theorem 4.8 with one slight change: Since disjunction is not allowed in DLV[not, $w$] programs, we replace the disjunctive fact $w \vee w'$. by two rules with (unstratified) negation: $w :\text{-} \text{ not } w'$. and $w' :\text{-} \text{ not } w$. Evidently, the meaning of the program remains unchanged, and we obtain a reduction from brave reasoning on DLV[not, $w$] programs to cautious reasoning for DLV[not, $w$]. □

Next, we point out the complexity of cautious reasoning over the "easy" fragments of the DLV language, where existence and uniqueness of an answer set is guaranteed.

THEOREM 4.12. *Given a* DLV[$not_s$, $w$] *program* $\mathcal{P}$, *and an atom $A$ as input, deciding whether $A$ is true in all answer sets of* $\mathcal{P}$ *is* P-*complete. Hardness holds even if* $\mathcal{P}$ *is a* DLV[] *program.*

PROOF. Each program in the fragments DLV[$not_s$, $w$] and DLV[$w$] has precisely one answer set. Consequently, on these fragments brave and cautious reasoning coincide, and the statement follows from the results on the complexity of brave reasoning derived in [Buccafurri et al. 2000]. (The introduction of priority levels for weak constraints does not affect these results, as priority levels can be easily converted to plain weights as we have seen in Section 2.2 and the beginning of Section 4.4.) □

Finally, we address the problem of answer set checking. In particular, we study the complexity of all fragments of the DLV language where weak constraints are allowed; for the other ones, such results are (often implicitly) contained in preliminary papers.

First we show an interesting correspondence between the problem of cautious reasoning for programs without weak constraints and the problem of answer set checking for the corresponding fragment where weak constraints may occur in the programs.

LEMMA 4.13. *Let* $\mathcal{P}$ *be a* DLV[$v$] *program and $q$ an atom. Then, there exists a* DLV[$v$, $w$] *program* $\mathcal{P}'$ *and a model $M$ for* $\mathcal{P}'$, *such that $M$ is an answer set for* $\mathcal{P}'$ *if and only if $q \notin M'$ for each answer set $M'$ of* $\mathcal{P}$. *Moreover,* $\mathcal{P}'$ *and $M$ are computable in polynomial time from* $\mathcal{P}$ *and $q$, and, if* $\mathcal{P}$ *belongs to* DLV[$v_h$], *then* $\mathcal{P}'$ *belongs to* DLV[$v_h$, $w$].

PROOF. Let $\bar{q}$, $\bar{q}'$, and $n\bar{q}$ be fresh atoms. Then, define the following program $\mathcal{P}'$:

$$\bar{q} \vee n\bar{q}.$$
$$\bar{q}' :\!-\ \bar{q}.$$
$$p :\!-\ \bar{q}. \qquad\qquad \text{for all}\ \ p \in B_{\mathcal{P}} - \{q\},$$
$$q \vee \bar{q}' :\!-\ B(r). \qquad \text{for all}\ \ r \in \mathcal{P}\ \ \text{such that}\ \ q \in H(r),$$
$$H(r) :\!-\ B(r). \qquad \text{for all}\ \ r \in \mathcal{P}\ \ \text{such that}\ \ q \notin H(r),$$
$$:\!\sim \bar{q}.\ [1:1]$$
$$:\!\sim n\bar{q}, \text{not}\ q.\ [1:1]$$

First, consider the interpretation $\widehat{M} = B_{\mathcal{P}'} - \{q, n\bar{q}\}$. Note that $\widehat{M}$ is an answer set for $Rules(\mathcal{P}')$, because it is clearly closed under $Rules(\mathcal{P}')$ and it is minimal. Indeed, $\widehat{M}$ is the only minimal closed interpretation for $\mathcal{P}'$ containing $\bar{q}$, and all closed interpretations not containing $\bar{q}$ must contain $n\bar{q}$ and are hence incomparable with $\widehat{M}$ (w.r.t. set inclusion). Moreover, observe that the cost of $\widehat{M}$ is $H^{\mathcal{P}'}(\widehat{M}) = 1$.

We show that $\widehat{M}$ is not an (optimal) answer set for $\mathcal{P}'$ if and only if there is some answer set $M'$ for $\mathcal{P}$ containing $q$.

Indeed, suppose that $\widehat{M}$ is not an optimal answer set for $\mathcal{P}'$. Then, there is an (optimal) answer set $M$ for $\mathcal{P}'$ having cost $H^{\mathcal{P}'}(M) = 0$, i.e., all weak constraints are satisfied. By the first one, we conclude that $\bar{q} \notin M$, which, by the first rule in $\mathcal{P}'$, implies that $n\bar{q} \in M$. Thus, by the second weak constraint, we have that $q \in M$. By the minimality of $M$, we furthermore obtain that $\bar{q}' \notin M$. Now it is easy to see that $M' = M - \{n\bar{q}\}$ is an answer set of $\mathcal{P}$ which contains $q$.

Conversely, consider any answer set $M'$ for $\mathcal{P}$ containing the atom $q$. It is easy to see that $M = M' \cup \{n\bar{q}\}$ is an answer set for $Rules(\mathcal{P}')$. Moreover, $H^{\mathcal{P}'}(M) = 0$; hence, $\widehat{M}$ is not an optimal answer set for $\mathcal{P}'$. This proves the claim.

Clearly, $\mathcal{P}'$ is computable in polynomial time from $\mathcal{P}$ and $q$. Moreover, if $\mathcal{P}$ is head-cycle free, then $\mathcal{P}'$ is head-cycle free, too. Indeed, a new "bad" cycle would imply that one of the fresh atoms $\bar{q}$, $\bar{q}'$, and $n\bar{q}$ already occured in the body of some rule and also in the head of some rule with nonempty body. Obviously, this is not the case. □

THEOREM 4.14. *Checking whether a given model $M$ is an answer set for a* DLV *program $\mathcal{P}$ is*

(1) P-*complete, if $\mathcal{P}$ belongs to* DLV[$\text{not}_s, w$]. *Hardness holds even if $\mathcal{P}$ is positive.*

(2) *co-NP-complete, if $\mathcal{P}$ belongs to* DLV[$\text{v}_\text{h}, \text{not}, w$]. *Hardness holds even if $\mathcal{P}$ is either positive or non-disjunctive.*

(3) $\Pi_2^P$-*complete, if $\mathcal{P}$ belongs to* DLV[$\text{v}, \text{not}, w$]. *Hardness holds even if $\mathcal{P}$ is positive.*

PROOF. 1). A DLV[$\text{not}_s, w$] program $\mathcal{P}$ has a unique answer set, computable in polynomial time. Thus, weak constraints do not affect the complexity of this problem, which remains the same as for DLV[$\text{not}_s$] and DLV[] programs, namely P-completeness. The hardness part for the latter is obtained by an easy reduction from deciding whether an atom $A$ is in the answer set of a DLV[] program $\mathcal{P}$: just add clauses $p :\!-\ A$. for all $p \in B_{\mathcal{P}}$ and check whether $B_{\mathcal{P}}$ is the answer set of the resulting program.

2). *Membership.* Let $\mathcal{P}$ be a program in DLV[$\text{v}_\text{h}, \text{not}, w$], and $M$ be a set of ground literals. We can check that $M$ is not an answer set for $\mathcal{P}$ as follows. First we verify in polynomial time whether $M$ is an answer set for $Rules(\mathcal{P})$. If this is not the case, we stop. Otherwise, we compute its cost $c = H^{\mathcal{P}}(M)$, and then decide whether there exists an answer set $M'$ for $\mathcal{P}$ such that $H^{\mathcal{P}}(M') < c$. From Lemma 4.9, this task is feasible in NP, and thus the checking problem is in co-NP.

*Hardness.* Recall that, given a DLV[$v_h$] program $\mathcal{P}$ and an atom $q$, deciding whether $q \notin M$ for each answer set of $\mathcal{P}$ is co-NP-complete, cf. [Eiter et al. 1998b]. From Lemma 4.13, this problem can be reduced to answer set checking for DLV[$v_h, w$] programs.

Moreover, it is well-known that, for each DLV[$v_h$] program $\mathcal{P}$, we can construct in polynomial time a DLV[not] program having the same answer sets as $\mathcal{P}$, by replacing disjunction by unstratified negation [Ben-Eliyahu and Dechter 1994]. The same reduction clearly allows us to reduce answer set checking for DLV[$v_h, w$] programs to answer set checking for DLV[not, $w$] programs. Thus, the latter problem is co-NP-hard, as well.

3). *Membership.* Let $\mathcal{P}$ be a program in DLV[$v$, not, $w$], and $M$ be a set of ground literals. We show that the complementary problem of checking that $M$ is not an answer set for $\mathcal{P}$ is in $\Sigma_2^P$. First we decide in co-NP whether $M$ is an answer set for $Rules(\mathcal{P})$ or not. If this is not the case, we stop. Otherwise, we compute its cost $c = H^{\mathcal{P}}(M)$ and then decide whether there exists an answer set $M'$ for $\mathcal{P}$ such that $H^{\mathcal{P}}(M') < c$. This is feasible in $\Sigma_2^P$, according to Lemma 4.6.

*Hardness.* Deciding whether a given literal is not contained in any answer set of a DLV[$v$] program is $\Pi_2^P$-complete [Eiter et al. 1997b]. From Lemma 4.13 it follows that this problem can be reduced to answer set checking for DLV[$v, w$] programs. □

## 4.5 Summary of Results and Discussion

|  | {} | {**w**} | {not$_s$} | {not$_s$, **w**} | {not} | {not, **w**} |
|---|---|---|---|---|---|---|
| {} | P | P | P | P | NP | $\Delta_2^P$ |
| {$v_h$} | NP | $\Delta_2^P$ | NP | $\Delta_2^P$ | NP | $\Delta_2^P$ |
| {$v$} | $\Sigma_2^P$ | $\Delta_3^P$ | $\Sigma_2^P$ | $\Delta_3^P$ | $\Sigma_2^P$ | $\Delta_3^P$ |

Table I.    The Complexity of Brave Reasoning in fragments of the DLV Language

|  | {} | {**w**} | {not$_s$} | {not$_s$, **w**} | {not} | {not, **w**} |
|---|---|---|---|---|---|---|
| {} | P | P | P | P | co-NP | $\Delta_2^P$ |
| {$v_h$} | co-NP | $\Delta_2^P$ | co-NP | $\Delta_2^P$ | co-NP | $\Delta_2^P$ |
| {$v$} | co-NP | $\Delta_3^P$ | $\Pi_2^P$ | $\Delta_3^P$ | $\Pi_2^P$ | $\Delta_3^P$ |

Table II.    The Complexity of Cautious Reasoning in fragments of the DLV Language

The complexity of Brave Reasoning and Cautious Reasoning from ground DLV programs are summarized in Table I and Table II, respectively. In Table III, we report both well-known (for the weak constraint-free case) and new results on the complexity of Answer Set Checking.

|        | $\{\}$ | $\{\mathbf{w}\}$ | $\{\mathrm{not}_s\}$ | $\{\mathrm{not}_s, \mathbf{w}\}$ | $\{\mathrm{not}\}$ | $\{\mathrm{not}, \mathbf{w}\}$ |
|--------|--------|------|---------|-------------|-------|------------|
| $\{\}$ | P | P | P | P | P | co-NP |
| $\{\mathrm{v_h}\}$ | P | co-NP | P | co-NP | P | co-NP |
| $\{\mathrm{v}\}$ | co-NP | $\Pi_2^P$ | co-NP | $\Pi_2^P$ | co-NP | $\Pi_2^P$ |

Table III.  The Complexity of Answer Set Checking in fragments of the DLV Language

The rows of the tables specify the form of disjunction allowed; in particular, $\{\}$ = no disjunction, $\{\mathrm{v_h}\}$ = head-cycle free disjunction, and $\{\mathrm{v}\}$ = unrestricted (possibly not head-cycle free) disjunction. The columns specify the support for negation and weak constraints. For instance, $\{w, \mathrm{not}_s\}$ denotes weak constraints and stratified negation. Each entry of the table provides the complexity of the corresponding fragment of the language, in terms of a completeness result. For instance, $(\{\mathrm{v_h}\}, \{\mathrm{not}_s\})$ is the fragment allowing head-cycle free disjunction and stratified negation, but no weak constraints. The corresponding entry in Table I, namely NP, expresses that brave reasoning for this fragment is NP-complete. The results reported in the tables represent completeness under polynomial time (and in fact LOGSPACE) reductions. All results have either been proved in Section 4.4 or emerge from [Eiter et al. 1997b; Gottlob 1994; Eiter et al. 1998b; Eiter and Gottlob 1995; Buccafurri et al. 2000]. Note that the presence of weights besides priority levels in weak constraints does not increase the complexity of the language, and thus the complexity results reported in [Buccafurri et al. 2000] remain valid also for our more general language. Furthermore, not all complexity results in the quoted papers were explicitly stated for LOGSPACE reductions, but can be easily seen to hold from (suitably adapted) proofs.

Looking at Table I, we see that limiting the form of disjunction and negation reduces the respective complexity. For disjunction-free programs, brave reasoning is polynomial on stratified negation, while it becomes NP-complete if we allow unrestricted (nonmonotonic) negation. Brave reasoning is NP-complete on head-cycle free programs even if no form of negation is allowed. The complexity jumps one level higher in the Polynomial Hierarchy, up to $\Sigma_2^P$-complexity, if full disjunction is allowed. Thus, disjunction seems to be harder than negation, since the full complexity is reached already on positive programs, even without any kind of negation. Weak constraints are irrelevant, from the complexity viewpoint, if the program has at most one answer set (if there is no disjunction and negation is stratified). On programs with multiple answer sets, weak constraints increase the complexity of reasoning moderately, from NP and $\Sigma_2^P$ to $\Delta_2^P$ and $\Delta_3^P$, respectively.

Table II contains results for cautious reasoning. One would expect its complexity to be symmetric to the complexity of brave reasoning, that is, whenever the complexity of a fragment is $C$ under brave reasoning, one expects its complexity to be co-$C$ under cautious reasoning (recall that co-P = P, co-$\Delta_2^P = \Delta_2^P$, co-$\Sigma_2^P = \Pi_2^P$, and co-$\Delta_3^P = \Delta_3^P$).

Surprisingly, there is one exception: While full disjunction raises the complexity of brave reasoning from NP to $\Sigma_2^P$, full disjunction alone is not sufficient to raise the complexity of cautious reasoning from co-NP to $\Pi_2^P$. Cautious reasoning remains in co-NP if default negation is disallowed. Intuitively, to disprove that an atom $A$ is a cautious consequence of a program $\mathcal{P}$, it is sufficient to find *any model* $M$ of $\mathcal{P}$ (which need not be an answer set or a minimal model) which does not contain $A$. For not-free programs, the existence of such a model guarantees the existence of a subset of $M$ which is an answer set of $\mathcal{P}$ (and does not contain $A$).

The complexity results for Answer Set Checking, reported in Table III, help us to understand the complexity of reasoning. Whenever Answer Set Checking for weak constraint-free programs is co-NP-complete for a fragment $F$, the complexity of brave reasoning jumps up to the second level of the Polynomial Hierarchy ($\Sigma_2^P$). In contrast, co-NP-completeness for Answer Set Checking involving weak constraints causes only a modest increase for brave reasoning, which stays within the same level ($\Delta_2^P$). Indeed, brave reasoning on full DLV programs suffers from three sources of complexity:

($s_1$)  the exponential number of answer set "candidates",

($s_2$)  the difficulty of checking whether a candidate $M$ is an answer set (the minimality of $M$ can be disproved by an exponential number of subsets of $M$), and

($s_3$)  the difficulty of determining the optimality of the answer set w.r.t. the violation of the weak constraints.

Now, disjunction (unrestricted or even head-cycle free) or unrestricted negation preserve the existence of source ($s_1$), while source ($s_2$) exists only if full disjunction is allowed (see Table III). Source ($s_3$) depends on the presence of weak constraints, but it is effective only in case of multiple answer sets (i.e., only if source ($s_1$) is present), otherwise it is irrelevant. As a consequence, e.g., the complexity of brave reasoning is the highest ($\Delta_3^P$) on the fragments preserving all three sources of complexity (where both full disjunction and weak constraints are allowed). Eliminating weak constraints (source ($s_3$)) from the full language, decreases the complexity to $\Sigma_2^P$. The complexity goes down to the first level of PH if source ($s_2$) is eliminated, and is in the class $\Delta_2^P$ or NP depending on the presence or absence of weak constraints (source ($s_3$)). Finally, avoiding source ($s_1$) the complexity falls down to P, as ($s_2$) is automatically eliminated, and ($s_3$) becomes irrelevant.

We close this section with briefly addressing the complexity and expressiveness of non-ground programs. A non-ground program $\mathcal{P}$ can be reduced, by naive instantiation, to a ground instance of the problem. The complexity of this ground instantiation is as described above. In the general case, where $\mathcal{P}$ is given in the input, the size of the grounding $Ground(\mathcal{P})$ is single exponential in the size of $\mathcal{P}$. Informally, the complexity of Brave Reasoning and Cautious Reasoning increases accordingly by one exponential, from P to EXPTIME, NP to NEXPTIME, $\Delta_2^P$ to EXPTIME$^{NP}$, $\Sigma_2^P$ to NEXPTIME$^{NP}$, etc. For disjunctive programs and certain fragments of the DLV language, complexity results in the non-ground case have been derived e.g. in [Eiter et al. 1997b; Eiter et al. 1998b]. For the other fragments, the results can be derived using complexity upgrading techniques [Eiter et al. 1997b; Gottlob et al. 1999]. Answer Set Checking, however, increases exponentially up to co-NEXPTIME$^{NP}$ only in the presence of weak constraints, while it stays in PH if no weak constraints occur. The reason is that in the latter case, the conditions of an answer set can be checked using small guesses, and no alternative (perhaps exponentially larger) answer set candidates need to be considered.

Finally, we remark that, viewed as a database-style query language (cf. [Dantsin et al. 2001]), the DLV language without weak constraints captures the classes of $\Sigma_2^P$ and $\Pi_2^P$ queries under brave and cautious reasoning, respectively, as follows from the results of [Eiter et al. 1997b]. DLV with weak constraints is more expressive, and captures the class of $\Delta_3^P$ queries. Thus, the full language of DLV is strictly more expressive than Disjunctive

Datalog (unless the polynomial hierarchy collapses). For instance, the Preferred Strategic Companies problem in Section 3.3.7 can be naturally expressed in the full DLV language, but it cannot be expressed at all if weak constraints are disallowed.

## 5. DLV FRONT-ENDS

Besides its kernel, the DLV system offers a number of front-ends for various domain specific reasoning formalisms, which we briefly overview in this section. For in-depth descriptions and download information for executables, we must refer to the quoted sources.

Currently, the DLV system has "internal" front-ends for inheritance reasoning, model-based diagnosis, planning, and SQL3 query processing; several "external" front-ends have been made available by other research teams. Each front-end maps its problem specific input into a DLV program, invokes the DLV kernel, and then post-processes any answer set returned, extracting from it the desired solution; optionally, further solutions are generated.

### 5.1 Internal Front-Ends

5.1.1 *Inheritance Front-End.* DLV's inheritance front-end supports an extension of the kernel language, named $\mathrm{DLP}^<$ [Buccafurri et al. 2002], in which rules can be grouped to objects arranged in a partial order $<$ (i.e., irreflexive and transitive relation), specified by the immediate successor relation ":". It mimics inheritance, where $o < o'$ reads "$o$ is more specific than $o'$", and assigns a "plausibility" to rules for conflict resolution. This is accomplished by overriding: informally, a rule $r$ is overridden, if the complement of every literal in the head of $r$ is supported by a more specific rule. The following simple example illustrates the approach; for formal details, we refer to [Buccafurri et al. 2002].

EXAMPLE 5.1. Consider the following program $\mathcal{P}_{tweety}$:

$$bird\,\{flies.\} \qquad penguin : bird\,\{\neg flies.\} \qquad tweety : penguin\,\{\,\}$$

It has the objects $bird$, $penguin$, and $tweety$, where $tweety$ is more specific than $penguin$ and $penguin$ more than $bird$; hence they are ordered by $tweety<penguin$, $penguin<bird$, and $tweety<bird$. The rules of each object are enclosed in "{" and "}".

$\mathcal{P}_{tweety}$ has the single answer set $\{\neg flies\}$, which fully captures its intuitive meaning. The rule $flies$ in object $bird$ is overridden by $\neg flies$ in the more specific object $penguin$.

Further information is available at `http://www.dlvsystem.com/inheritance/`.

5.1.2 *Diagnosis Front-End.* In model-based diagnosis, two major formal notions of diagnosis have been proposed: *abductive diagnosis* [Poole 1989] and *consistency-based diagnosis* [Reiter 1987]. Both strive for capturing, given a logical background theory $T$, some observations $O$, and a set of hypotheses $H$ on potential failures, a diagnosis $\Delta \subseteq H$ which reconciles $T$ with $O$ (i.e., $T \cup H$ models $O$ resp. $T \cup H \cup O$ is consistent).

DLV's built-in diagnosis front-end [Eiter et al. 1997a] supports, under certain restrictions, an instance of abductive diagnosis for $T$ being a logic program and consistency-based diagnosis for $T$ under classical semantics. Moreover, it offers computation of different types of diagnoses, viz. general, $\subseteq$-minimal (*irredundant*), and single failure diagnoses. The following simple example illustrates the usage of logic program semantics.

EXAMPLE 5.2 NETWORK DIAGNOSIS. Suppose in the computer network of Figure 1 machine $a$ is up, but we observe that it cannot reach machine $e$. Which machines are down?

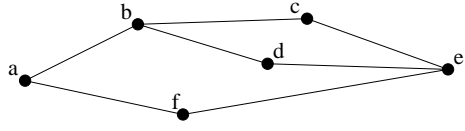

Fig. 1.   Computer network

This can be easily modeled as a diagnosis problem, where the theory $T$ is the logic program

$$reaches(X, X) \coloneq node(X), \mathrm{not}\ down(X).$$
$$reaches(X, Z) \coloneq reaches(X, Y), connected(Y, Z), \mathrm{not}\ down(Z).$$

the observations $O$ are $\mathrm{not}\ down(a)$ and $\mathrm{not}\ reaches(a, e)$, and the hypotheses $H$ are $down(a), \ldots, down(f)$. Running the diagnosis front-end, we can compute all diagnoses; the irredundant ones are $\{down(e)\}$, $\{down(b),\ down(f)\}$, and $\{down(c),\ down(d), down(f)\}$. Note that a logic programming representation of $T$ is beneficial, since non-reachability is easily expressed; under classical semantics, this is more cumbersome.

5.1.3 *Planning Front-End.* DLV's has a powerful built-in planning front-end called DLV$^{\mathcal{K}}$ [Eiter et al. 2003a]. It is based on the action language $\mathcal{K}$ [Eiter et al. 2000b; 2001b], which is akin to the action language $\mathcal{C}$ [Giunchiglia and Lifschitz 1998] but semantically adheres to logic programming instead of classical logic. It features dealing with incomplete knowledge (neither an atom $f$ nor $\neg f$ may be known in a state), negation as failure, nondeterministic action effects, and parallel actions. A DLV$^{\mathcal{K}}$ program describes an initial state, the state transitions by action executions, and the goal to be reached. Different kinds of plans may be computed; among them are *optimal plans* [Eiter et al. 2002b], in which the actions have minimal total cost, according to specified action costs.

EXAMPLE 5.3. The Traveling Salesperson Problem in Section 3.3.3 (with input facts *node*, *arc*, and *start*) is expressed by the following DLV$^{\mathcal{K}}$ program:

$\quad fluents:\quad in(C)\ requires\ node(C).$
$\qquad\qquad\quad\ visited(C)\ requires\ node(C).$
$\qquad\qquad\quad\ unvisited.\ home.$
$\quad actions:\quad travel(X, Y)\ requires\ arc(X, Y, C)\ costs\ \ C.$
$\quad always:\quad\ executable\ \ travel(X, Y)\ \ if\ \ in(X).$
$\qquad\qquad\quad\ nonexecutable\ \ travel(X, Y)\ \ if\ \ visited(Y).$
$\qquad\qquad\quad\ caused\ \ in(Y)\ \ after\ \ travel(X, Y).$
$\qquad\qquad\quad\ caused\ \ visited(Y)\ \ after\ \ travel(X, Y).$
$\qquad\qquad\quad\ caused\ \ unvisited\ \ if\ \ node(C),\ \ not\ \ visited(C).$
$\qquad\qquad\quad\ caused\ \ home\ \ if\ \ in(C), start(C).$
$\qquad\qquad\quad\ inertial\ \ visited(C).$
$\qquad\qquad\quad\ noConcurrency.$
$\quad initially:\quad caused\ \ in(C)\ \ if\ \ start(C).$
$\quad goal:\qquad\ not\ \ unvisited,\ home\ ?\ (\langle n \rangle)$

Informally, the *fluents* and *actions* parts declare changeable predicates (*fluents*) and actions, respectively, where after *requires* argument types and action costs are specified. The *always* part describes when action *travel* can be taken, as well as causal laws for fluents, depending on other predicates in the current and previous state and possible actions executed; the law in the *initially* part applies only to the initial state. The final *goal* part defines the goal by two ground literals and a plan length (here the number of nodes, $\langle n \rangle$). More information and examples are available at `http://www.dlvsystem.com/K/`.

5.1.4 *SQL3 Front-End.* Different from earlier versions, SQL3 allows recursive database queries. DLV's built-in SQL3 front-end, available since the early days, provides a playground for getting a grasp on SQL3, which formerly database systems did not support. It covers an SQL3 fragment, including queries like the "list-of-materials" query.

5.1.5 *Meta-Interpreter Front-End.* A meta-interpretation technique for prioritized logic programs is described in [Eiter et al. 2001a; 2003], developed for computing different notions of preferred answer sets from [Brewka and Eiter 1999; Schaub and Wang 2001]. In this setting, a particular fixed program, the meta-interpreter, is used in combination with an input prioritized logic program, which is represented by a set of facts. The answer sets of the meta-interpreter augmented with the facts then correspond to the preferred answer sets of the original prioritized program. The meta-interpreters, several examples, and further information are available at `http://www.dlvsystem.com/preferred/`.

## 5.2 External Front-Ends

*The* plp *Front-End.* The *plp* system for prioritized logic programs [Delgrande et al. 2001], available at `http://www.cs.uni-potsdam.de/~torsten/plp/`, is the most notable external front-end. It is a powerful platform for declaratively "programming" preference-semantics for logic programs, by respecting particular orders and criteria for rule consideration. *plp* can use DLV but also Smodels as a back-end for computation.

*Update Front-End.* A front-end for update logic programs, i.e., sequences $\mathbf{P} = (P_0, \ldots, P_n)$ of extended logic programs $P_i$, where $P_0$ is the initial program and $P_i$, $i > 0$, represents an update at time $i \in \{1, 2, \ldots, n\}$, is described in [Eiter et al. 2001e]. The semantics of $\mathbf{P}$, defined in [Eiter et al. 2002c], extends standard answer sets for a single program ($n = 1$). The front-end supports different reasoning tasks and was used to realize policies for updating knowledge bases [Eiter et al. 2001d]; it is available for download at `http://www.kr.tuwien.ac.at/staff/giuliana/project.html`.

*The* nlp *Front-End.* Most recent is the *nlp* front-end, which transforms nested logic programs to Answer Set Programs; see [Pearce et al. 2002] for a description and `http://www.cs.uni-potsdam.de/~torsten/nlp/` for a prototype download.

## 6. THE IMPLEMENTATION OF THE DLV SYSTEM: AN OVERVIEW

The high expressiveness of the DLV language together with the ambition to deal efficiently also with larger instances of simpler problems, made the implementation of DLV system a challenging task.

The DLV core has three layers (see Figure 2), each of which is a powerful subsystem per se: The *Intelligent Grounder* (IG, also *Instantiator*) has the power of a deductive database system; the *Model Generator* (MG) is as powerful as a Satisfiability Checker; and the *Model Checker* (MC) is capable of solving co-NP-complete problems. These layers have been improved over the years through the implementation of sophisticated data structures and of advanced optimization techniques. It is infeasible to give a detailed description of the implementation of DLV in this paper. Instead, we describe the general architecture of DLV, illustrate the inspiring principles underlying the DLV system, and give an overview of the main techniques which were employed in the implementation, pointing the reader to specific papers that provide full details.

The system architecture of DLV is shown in Figure 2. The internal system language is the one described in Section 2, and the DLV Core (the shaded part of the figure) is an efficient engine for computing answer sets (one, some, or all) of its input. In addition to various front-ends (described in Section 5), there is a Graphical User Interface (GUI) that provides convenient access to some of these front-ends as well as the system itself.

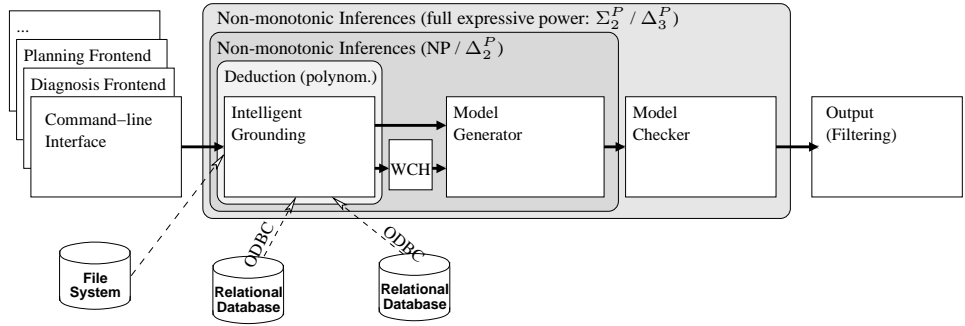

Fig. 2.    The System Architecture of DLV

The implementation of the DLV system is based on very solid theoretical foundations, and exploits the results on the computational complexity of DLV language fragments discussed in Section 4. Ideally, the performance of a system should reflect the complexity of the problem at hand, such that "easy" problems (say, those of polynomial complexity) are solved fast, while only harder problems involve methods of higher run-time cost. Indeed, the DLV system is designed according to this idea, and thrives to exploit the complexity results reported in Section 4. Note that in general, without syntactic restrictions, it is impossible to detect whether a program uniformly encodes an "easy" problem, since this task is clearly undecidable.

For example, stratified normal programs (which have polynomial complexity, as reported in Table I) are evaluated solely using techniques from the field of deductive databases, without employing the more complex techniques which are needed to evaluate full DLV programs; in fact, such normal stratified programs are evaluated without generating the program instantiation at all. Datalog programs encoding deductive problems like Reachability or Same Generation (cf. Sections 3.1.1 and 3.1.2) and SQL3 front-end, which only generates such programs, benefit from this and exhibit good performance.

The architecture of the DLV Core closely reflects complexity results for various subsets of our language. As mentioned before, the Intelligent Grounding (IG) module is able to completely solve some problems which are known to be of polynomial time complexity (like normal stratified programs); the Model Generator (together with the Grounding) is capable of solving NP-complete problems. Adding the Model Checker is needed to solve $\Sigma_2^P$-complete problems. The WCH (Weak Constraints Handler) comes into play only in presence of weak constraints. More precisely, referring to the notation of Section 4, we have the following five disjoint language classes $L_1 - L_5$ for evaluation:

—$L_1$ contains the programs included in the class $\langle \{\}, \{w, \mathrm{not}_s\} \rangle$, which all have polynomial complexity. They are completely evaluated by the IG module,[8] which runs in polynomial time (referring to propositional complexity).

---

[8]In this section, for *evaluation* we mean the process of generating an answer set of the input program; brave reasoning corresponds to this task.

—$L_2$ contains the programs which are in the subclass corresponding to $\langle\{\,\mathtt{v_h}\,\},\{\mathrm{not}\}\rangle$, but not in $L_1$. The complexity of this fragment is NP, and the programs are evaluated by the MG (besides the IG) with only a call to the linear-time part of the MC module. Note that the MG implements a flat backtracking algorithm and is suitable for solving NP-complete problems.

—$L_3$ contains the DLV programs from $\langle\{\,\mathtt{v_h}\,\},\{\mathrm{not},w\}\rangle$ minus $L_1 \cup L_2$. The complexity of this fragment is $\Delta_2^P$. Here, also the WCH module is employed, which iteratively invokes the MG. Again, only the linear-time part of the MC is invoked.

—$L_4$ contains the programs from the subclass corresponding to $\langle\{\,\mathtt{v}\,\},\{\mathrm{not}\}\rangle$ minus $L_1 \cup L_2 \cup L_3$. The complexity of this fragment is $\Sigma_2^P$, and the programs are evaluated by the MG module (besides the IG) with calls to the *full* MC module. Note that a flat backtracking algorithm is not sufficient to evaluate $\Sigma_2^P$-complete problems, and such a nested evaluation scheme, with calls to MC, is needed.

—Finally, $L_5$ contains all other programs, i.e., those in the full class (corresponding to $\langle\{\,\mathtt{v}\,\},\{\mathrm{not},w\}\rangle$) which are not contained in $L_1 \cup L_2 \cup L_3 \cup L_4$, where we have the full language complexity of $\Delta_3^P$. The evaluation proceeds as for $L_4$, but also the WCH module comes into play for handling the weak constraints.

The three DLV modules, MG, MC, and WCH, thus deal with the three sources of complexity denoted by $(s_1)$, $(s_2)$, and $(s_3)$ in Section 4; each of them is fully activated *only if* the respective source of complexity is present in the program at hand.

Let us now look at the evaluation flow of the DLV computation in some more detail. Upon startup, the DLV Core or one of the front-ends parses the input specified by the user and transforms it into the internal data structures of DLV. In both cases, this is done efficiently (requiring only linear memory and time). The input is usually read from text files, but DLV also provides a bridge to relational databases through an ODBC interface, which allows for retrieving facts stored in relational tables.

Using differential and other advanced database techniques (see [Faber et al. 1999; Leone et al. 2001]) together with suitable data structures, the *Intelligent Grounding* (IG) module then efficiently generates a ground instantiation $Ground(\mathcal{P})$ of the input that has the same answer sets as the full program instantiation, but is much smaller in general. For example, in case of a stratified program, the IG module already computes the single answer set, and does not produce any instantiation.

The heart of the computation is then performed by the Model Generator and the Model Checker. Roughly, the former produces some "candidate" answer sets, the stability of which is subsequently verified by the latter. In presence of weak constraints, further processing is needed, which is performed under the control of the WCH module. Since the handling of weak constraints is somehow orthogonal to the rest of the computation, we first focus on the evaluation of standard disjunctive logic programs, describing the processing of weak constraints later on.

The generation of the answer sets of a program $\mathcal{P}$ relies on a monotonic operator $\mathcal{W}_\mathcal{P}$ [Leone et al. 1997] which extends the well-founded operator of [van Gelder et al. 1991] for normal programs to disjunctive programs. It is defined in terms of a suitable notion of *unfounded set*. Intuitively, an unfounded set for a disjunctive program $\mathcal{P}$ w.r.t. an interpretation $I$ is a set of positive literals that cannot be derived from $\mathcal{P}$ assuming the facts in $I$ [Leone et al. 1997].

Briefly, the MG works as follows: First, $\mathcal{W}_\mathcal{P}^\omega(\emptyset)$ (the fixpoint of $\mathcal{W}_\mathcal{P}$) is computed, which is contained in every answer set. If $\mathcal{W}_\mathcal{P}^\omega(\emptyset)$ is a total model, it is returned as the (unique) answer set. Otherwise, moving from $\mathcal{W}_\mathcal{P}^\omega(\emptyset)$ towards the answer sets, a literal (called *possibly-true literal* in [Leone et al. 1997]), the truth of which allows to infer new atoms, is assumed true. Clearly the choice of "good" possibly-true literals at each step (i.e., a sequence of possibly-true literals that quickly leads to an answer set) is crucial for an efficient computation, so we employ novel heuristics with extensive lookahead and also propagate knowledge about choices that lead to inconsistency [Faber et al. 2001].

The computation proceeds by alternately selecting a possibly-true literal and applying the pruning operator, until either a total model of $Ground(\mathcal{P})$ is reached or two contradictory literals are derived. If a model is found, the Model Checker is called; otherwise, backtracking is performed.

The *Model Checker* (MC) verifies whether the model $M$ at hand is an answer set for the input program $\mathcal{P}$. In particular, the MC disregards weak constraints, and verifies whether $M$ is an answer set for $Rules(\mathcal{P})$; the optimality of the models w.r.t. the violation of weak constraints is handled by the WCH module. The task performed by MC is very hard in general, because checking the stability of a model is well-known to be co-NP-complete (cf. [Eiter et al. 1997b]). However, for some relevant and frequently used classes of programs answer-set checking can be efficiently performed (see Table III in Section 4).

The MC implements novel techniques for answer-set checking [Koch and Leone 1999], which extend and complement previous results [Ben-Eliyahu and Dechter 1994; Ben-Eliyahu and Palopoli 1994; Leone et al. 1997]. The MC fully complies with the complexity bounds specified in Section 4. Indeed, (a) it terminates in polynomial time on every program where answer-set checking is tractable according to Table III (including, e.g., HCF programs); and (b) it always runs in polynomial space and single exponential time. Moreover, even on general (non-HCF) programs, the MC limits the inefficient part of the computation to the modules that are not HCF. Note that it may well happen that only a very small part of the program is not HCF [Koch and Leone 1999].

Finally, once an answer set has been found, the control is returned to the front-end in use, which performs post-processing and possibly invokes the MG to look for further models. In a sense, also the language described in Section 2 is implemented by means of a front-end, even though by a very "thin" one.

In presence of weak constraints, after the instantiation of the program, the computation is governed by the WCH and consists of two phases: (i) the first phase determines the cost of an optimal answer set[9], together with one "witnessing" optimal answer set and, (ii) the second phase computes all answer sets having that optimal cost. It is worthwhile noting that both the IG and the MG also have built-in support for weak constraints, which is activated (and therefore incurs higher computational cost) only if weak constraints are present in the input. The MC, instead, does not need to provide any support for weak constraints, since these do not affect answer-set checking at all.

---

[9]By *cost* of an answer set we mean the sum of the weights of the weak constraints violated by the answer set, weighted according to their priority level – see Section 2.2.

## 7. EXPERIMENTS AND BENCHMARKS

### 7.1 Overview of Compared Systems

The goal of the DLV project is to provide an efficient implementation of Disjunctive Logic Programming. Thus, the main purpose of the experiments is to assess the efficiency of DLV as a DLP system. Unfortunately, most DLP systems in the literature (cf. Introduction) have been built for experimental purposes only and are not much concerned with efficiency; there are almost no elaborated, robust implementations of DLP to evaluate the performance of DLV against. Only the GnT system [Janhunen et al. 2000; Janhunen et al. 2003] (described below) is an efficient and robust implementation of DLP under the answer set semantics, which we thus selected for an experimental comparison with DLV.

Furthermore, in order to elucidate the performance of DLV in comparison with more specialized systems, we have also considered some efficient ASP engines which are close relatives of DLV, but do not support disjunction. The comparison of DLV with these systems is of other interest than its comparison with DLP systems: it should provide an idea how disjunctive systems like DLV and GnT compare (on natural disjunctive encodings) against non-disjunctive systems which use unstratified negation. It is worthwhile noting, however, that non-disjunctive ASP systems have lower expressiveness than DLV. In particular, on benchmark problems in NP, DLV (and GnT as well) has an overhead with respect to these systems since it is designed as a solver for a larger class containing much harder problems, while non-disjunctive systems are tailored for a special fragment in this class which is, in general, not efficiently recognizable in a syntactic way.

Recently, the scientific community has made considerable efforts on the implementation of answer set programming, and a number of systems nowadays support this formalism to some extent [Anger et al. 2001; Aravindan et al. 1997; Babovich 2002; Chen and Warren 1996; Cholewiński et al. 1996; Cholewiński et al. 1999; East and Truszczyński 2000; East and Truszczyński 2001a; East and Truszczyński 2001b; Egly et al. 2000; Eiter et al. 1998a; Janhunen et al. 2000; Janhunen et al. 2003; Lin and Zhao 2002; McCain and Turner 1998; Niemelä and Simons 1997; Rao et al. 1997; Seipel and Thöne 1994; Simons et al. 2002]. Most of these systems have common roots, but differ in several respects, especially in the languages they support.

For a comparison with DLP systems, we have considered ASP systems whose languages are as close to Disjunctive Logic Programming (and to the DLV language) as possible, since we feel that it is difficult to make a fair comparison among systems supporting very different languages. We have thus focused on systems supporting the full language of function-free logic programs under the answer set semantics, including recursion and unstratified negation. Among those, we preferred more widely used systems which are at a more advanced engineering stage; several other ASP systems are research prototypes implemented just for experimental purposes, and would need quite some engineering work to be made efficient and robust.

In particular, in addition to GnT, we picked *Smodels* [Simons et al. 2002; Niemelä et al. 2000; Niemelä and Simons 1997] – one of the most robust and widely used ASP systems – and *ASSAT* [Lin and Zhao 2002], which is not as widely used and robust as Smodels yet, but appeared to be very efficient in a couple of recent experimental comparisons [Simons et al. 2002; Lin and Zhao 2002].

It is worthwhile noting that another system named Cmodels [Babovich 2002] has recently emerged. Cmodels is similar in spirit to ASSAT, in that it translates the logic program to a CNF theory and uses a SAT-solver for its evaluation. The original version of Cmodels required the input program to satisfy a so called tightness condition, and it seems that this only applies to a single of the benchmarks which is not already solved by the instantiation procedure (RAMSEY). After completion of all our benchmarks, an extended version called Cmodels-2 became available, which is capable of handling arbitrary non-disjunctive programs, by implementing the same techniques as ASSAT. We did not consider Cmodels in our comparisons because the focus of this paper is on systems implementing the full language (including disjunction); we include the non-disjunctive systems Smodels and ASSAT mainly for reference purposes, and (historically) due to the limitations of the original version. Furthermore, since Cmodels-2 is based on similar ideas as ASSAT, we might suspect that it shows similar behavior. The standing of this system remains to be explored in other work providing an exhaustive comparison of answer set solvers, which is beyond the scope of this paper.

In the rest of this section, we provide a short description of the three systems we used for our comparison with to the "May 16th, 2003" release of DLV.

*Smodels.* [Simons et al. 2002; Niemelä et al. 2000; Niemelä and Simons 1997] is one of the best known and most widely used Answer Set Programming systems. It implements the answer set semantics for normal logic programs extended by built-in functions as well as cardinality and weight constraints for domain-restricted programs.

Disregarding the extension for cardinality and weight constraints, the Smodels system takes as input normal (v-free) rules as described in Section 2. Programs must be *domain-restricted*, which intuitively amounts to the following property: the predicate symbols in the program are divided into two classes, namely *domain predicates* and *non-domain* predicates, where the former are predicates whose definition[10] does not involve rules with negative recursion [Syrjänen 2002]. Every rule must be domain-restricted in the sense that every variable in that rule must appear in a domain predicate which appears positively in the rule body.

In addition to normal rules, Smodels supports rules with cardinality and weight constraints. To illustrate the idea on an example, the cardinality constraint

$$1\{a, b, not\, c\}2.$$

holds in an answer set if at least 1 but at most 2 of the literals in the constraint are satisfied by that answer set.

The instantiation module of Smodels is a separate application called *Lparse,* which preprocesses the programs which are then evaluated by Smodels.

In our tests, we used the current versions of these two at the time of this writing, Lparse 1.0.11 and Smodels 2.27, running them in a UNIX pipe and measuring overall CPU time.

*GnT.* [Janhunen et al. 2000; Janhunen et al. 2003] is a DLP system that extends the Smodels language by disjunction rule heads, written as $\{a_1, \cdots, a_n\}$. GnT has been implemented on top of Smodels by means of a rewriting technique. Roughly, a disjunctive input program $\mathcal{P}$ is rewritten to a non-disjunctive program $\mathcal{P}'$, such that the answer sets of $\mathcal{P}'$ form a superset of the answer sets of $\mathcal{P}$. Program $\mathcal{P}'$ is passed to Smodels for the evaluation, and each answer set $M$ produced is then processed by a nested call to Smodels, to check whether $M$ is also an answer set of $\mathcal{P}$. GnT is not a *strict* generalization of Smodels in that it does not support the full language of Smodels (in particular, it has no cardinality constraints).

---

[10]The definition of a predicate $p$ contains all rules with head $p$ plus the definitions of all predicates appearing in the bodies of the rules with head $p$.

For our tests, we used GnT 2 as downloaded from `http://www.tcs.hut.fi/Software/gnt/` in April 2003, together with Lparse 1.0.11, running both in a UNIX pipe and measuring overall CPU time.

*ASSAT.* The ASSAT system (Answer Sets by SAT solvers) is a system for computing answer sets of a logic program by using SAT solvers [Lin and Zhao 2002; Zhao 2002]. Given a ground logic program $\mathcal{P}$ and a SAT solver $X$, ASSAT$(X)$ works as follows:

—It computes the Clark-completion [Clark 1978] of $\mathcal{P}$ and converts it into a set $C$ of clauses.
—It then repeats the following steps:
   —Call $X$ on $C$ to get a model $M$ of $\mathcal{P}$, and terminate with failure if no such $M$ exists.
   —If $M$ is an answer set of $\mathcal{P}$, then return it as the result.
   —Otherwise, find some loops in $\mathcal{P}$ whose "loop formulas" (defined in [Lin and Zhao 2002]) are not satisfied by $M$, and add their clausal forms to $C$. This exploits the result that a model $M$ is not an answer set of $\mathcal{P}$ if and only if some loop formula is not satisfied [Lin and Zhao 2002].

As shown in [Lin and Zhao 2002], this procedure is sound and complete, assuming that $X$ is a sound and complete SAT solver. It should be noted that this approach is geared towards computing *one* answer set, rather than computing *all* (or a given number of) answer sets. While in an incremental computation, already computed answer sets can be easily excluded from re-computation by adding suitable clauses, the efficiency of this approach remains to be seen.

Like Smodels, ASSAT employs Lparse for program instantiation. It accepts the core language of Smodels without cardinality and weight constraints, and imposes the domain restriction constraint on the variables required by Lparse.

For the benchmarks, we used ASSAT version 1.34 together with Lparse 1.0.11 and the SAT solver Chaff version 2001-06-16, which we obtained from `http://www.ee.princeton.edu/~chaff/index1.html`. (By using Chaff, ASSAT provided the best performance in benchmarks by the authors of ASSAT [Lin and Zhao 2002].)

We used ASSAT in its default (built-in) configuration and Chaff with the default configuration file "cherry.smj," which comes with the Chaff distribution.

*Using the* DLV *instantiator with Smodels, GnT, and ASSAT.* In addition to the "native" setting where Smodels, GnT, and ASSAT all employ Lparse to compute the ground instantiation of their input, we also used the instantiator mode of DLV (requested by the command-line option `-instantiate`), where DLV just computes and prints the ground instantiation of its input which then can be fed to other systems. The actual invocation chain in this case was

```
dlv -instantiate | lparse | smodels/gnt2/assat
```

since Smodels, GnT, and ASSAT expect their input in a particular internal format only supported by Lparse (which performs very efficiently when only used as a filter to convert ground programs into that format, so in general the extra time employed by Lparse here is negligible).

## 7.2   Benchmark Problems and Data

We had to ponder about the construction of a suite of benchmark problems to test DLP systems, as unlike other areas, such as Satisfiability Testing, there currently is no generally recognized set of benchmark problems for DLP systems. We believe that a benchmark suite should be ample, cover different application areas, and comprise problems of different complexities, in order to highlight the suitability of the systems for different applications, and on different problem (complexity) classes.

We have studied previous experimentation work performed for benchmarking ASP systems at Helsinki University of Technology, and exploited also the precious discussions on this topic at *LPNMR'01*, the *AAAI Spring 2001 Symposium on Answer-Set Programming*, and the *Dagstuhl Seminar on Answer Set Programming* in September 2002.

We considered a number of benchmark problems taken from various domains and with very different complexities ranging from P over NP, co-NP, and $\Delta_2^P$ to $\Sigma_2^P$ and included most examples provided in Section 3 as well as a couple of further problems with suitable sets of benchmark data. In particular, we evaluated the systems on the following problems:

—Reachability (REACH)

—Same Generation (SAMEGEN)

—Hamiltonian Path (HAMPATH)

—Traveling Salesperson (TSP)

—Ramsey Numbers (RAMSEY)

—Sokoban (SOKOBAN)

—Quantified Boolean Formulas (2QBF)

—Strategic Companies (STRATCOMP)

All problems except TSP are from the suite of "Benchmark Problems for Answer Set Programming Systems" of the University of Kentucky (`http://cs.engr.uky.edu/ai/benchmarks.html`), which was used to compare ASP systems at the *6th International Conference on Logic Programming and Non-Monotonic Reasoning (LPNMR'01)* in Vienna, Sept. 2001. RAMSEY was also used to compare ASP systems at the *AAAI Spring 2001 Symposium on Answer Set Programming* in Stanford, March 2001, and SAMEGEN, HAMPATH, RAMSEY, and STRATCOMP are also in the benchmark suite defined at the *Dagstuhl Seminar on Answer Set Programming*, Sept. 2002 [Brewka et al. 2002].

As far as the encodings are concerned, we did not look for tricky ones; rather, we tried to exploit the expressiveness of the language to design simple and declarative encodings. We kept the encodings as close as possible to the standard DLP language,[11] avoiding system-specific language features.[12] In the main comparison (DLV vs GnT), we run the two systems precisely on the same (disjunctive) encoding, when using DLV instantiator for GnT; also when GnT was run with his native instantiator Lparse, we nearly used the same encodings for the two systems (only some domain predicate had to be added for GnT in some case). Disregarding domain predicates, also for Smodels and ASSAT we could use the same encodings adopted for DLV and GnT, by just translating disjunction to unstratified negation in the straightforward way (Smodels and ASSAT were not benchmarked on 2QBF and STRATCOMP, since their languages miss the needed expressiveness).

---

[11]The language adopted in [Gelfond and Lifschitz 1991] is generally acknowledged in the literature, and most systems support it to a large extent.

[12]Only for TSP we resorted to optimization constructs which are not included in standard DLP.

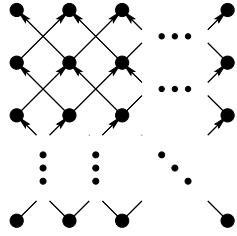

Fig. 3. Input for SAMEGEN

All benchmark instances and encodings we have used are available on the web at `http://www.dlvsystem.com/examples/tocl-dlv.zip`.

We next specify the data and the encodings which we have used for the experiments.

*Reachability (REACH).* For DLV we used the encoding presented in Section 3.1.1; for GnT, Smodels,[13] and ASSAT we had to make some modifications to respect the *domain restriction* constraint required by Lparse:

$$reachable(X,Y) :\!\!- arc(X,Y).$$
$$reachable(X,Y) :\!\!- arc(X,U), reachable(U,Y), vertex(Y).$$

In particular, we had to add $vertex(Y)$ to the body of the second rule in order to restrict the domain of the variable $Y$.

The input graphs for Reachability were generated by means of the Stanford GraphBase [Knuth 1994], using the function *random_graph(#nodes, #arcs, 0,0,0,0,0,0,0,0)* with a ratio of 3:1 between #arcs and #nodes.

*Same Generation (SAMEGEN).* Here, we used the encoding described in Section 3.1.2 and instances where the parent relation is a square board as depicted in Figure 3.

We did not need to adapt the encodings for GnT, Smodels, and ASSAT, as the encoding from Section 3.1.2 already satisfies the domain restriction constraint of Lparse.

*Ramsey Numbers (RAMSEY).* We considered deciding, for varying $k$, $m$, and $n$, if $n$ is the Ramsey number $R(k,m)$. For DLV, we used the encoding shown in Section 3.3.4.

It is worthwhile noting that for this problem we do not use a single, uniform encoding to solve all instances, rather, we have one program for each instance. In particular, for checking that $n$ is the Ramsey number $R(k,m)$, the first constraint contains $\binom{k}{2}$ atoms with predicate $red$ and the second constraint contains $\binom{m}{2}$ atoms with predicate $blue$.

For Smodels and ASSAT we replaced the disjunctive rule by two normal rules using unstratified negation. We also added some instances of $node(\dots)$ to the integrity constraints to meet the syntactical restrictions of Lparse . For these systems, we have also considered an alternative encoding where the domain restriction constraint is satisfied by adding an atom of the form $arc(t_1, t_2)$, for each atom $blue(t_1, t_2)$ and each atom $red(t_1, t_2)$ occur-ring in the constraint. This encoding seems to have a positive impact on the performance (see Section 7.3) due to the instantiation procedure employed by Lparse. However, we believe that this encoding is less intuitive, as a domain predicate is understood as the "type" of a variable; accordingly, $node(X)$ would be the natural type for each variable $X$ occurring in this program.

---

[13]Note that GnT behaves like Smodels if the input program can be solved by the instantiator.

For GnT, we used the disjunctive encoding from Section 3.3.4 with the addition of the above domain predicates to satisfy the domain restriction constraint.

*Hamiltonian Path (HAMPATH).* For DLV, we used the encoding described in Section 3.3.2. For Smodels and ASSAT we rewrote the disjunctive guessing rule to use unstratified negation instead, and added some instances of $arc(\dots)$ to the integrity constraints to work around the *domain restriction* constraint required by Lparse . [14]

For GnT we used the (disjunctive) encoding of DLV, with the addition of the domain predicates in the integrity constraints as above.

The graph instances were generated using a tool by Patrik Simons (`http://tcs.hut.fi/Software/smodels/misc/hamilton.tar.gz`) which originally was used to compare Smodels against SAT solvers [Simons 2000]. This tool generates graphs with nodes labeled $0$, $1$, and so forth, and we assume $node(0)$ as the starting node.

*Traveling Salesperson (TSP).* Here we have reused the graph instances for HAMPATH described above and randomly added costs between 1 and 10 to every arc. The uniform approach described in Section 3.3.3 for encoding TSP in DLV is not feasible using the language of Smodels, since the direct counterpart of the weak constraint of DLV in Smodels (the $minimize$ statement) does not support the use of variables. Thus, we employed a more direct, instance-specific encoding in the form of

$$minimize\ [\ inPath(0,3) = 9,\ inPath(0,2) = 4,\ \dots,\ inPath(3,1) = 7\ ].$$
$$compute\ all\ \{\}.$$

for Smodels and

$$:\sim inPath(0,3).[9:]$$
$$:\sim inPath(0,2).[4:]$$
$$\vdots$$
$$:\sim inPath(3,1).[7:]$$

for DLV (for the sake of similarity). The rest of the encoding for DLV corresponds to the one described in Section 3.3.3. For Smodels, we have replaced the disjunctive rules with rules employing unstratified negation.

Note that the $compute\ all$ statement for Smodels only leads to a single model of least cost, while during the computation Smodels also prints non-optimal models. To the best of our knowledge, Smodels currently is not capable of generating precisely the set of models having least cost (in case there is more than one such model). Such models can only be obtained by running Smodels a second time with the original program extended by a further constraint discarding the models whose costs are higher than the least cost (computed in the first run).

---

[14] We have used the domain predicate $arc$ instead of $node$, since Lparse seems to produce a more advantageous instantiation this way; cf. the discussion on the encodings of RAMSEY.

ASSAT and GnT currently do not support the *minimize* statement of Lparse/Smodels (and weak constraints neither), so we could not test them on this problem.

*Sokoban (SOKOBAN).* Sokoban means "warehouse-keeper" in Japanese, and is the name of a game puzzle which has been developed by the Japanese company Thinking Rabbit, Inc., in 1982. Each puzzle consists of a room layout (a number of square fields representing walls or parts of the floor, some of which are marked as storage space) and a starting situation (one sokoban and a number of boxes, all of which must reside on some floor location). The goal is to move all boxes to storage locations. To this end, the sokoban can walk on floor locations (unless occupied by a box), and push single boxes to unoccupied floor locations. Figure 4 shows a typical configuration involving two boxes, where grey fields are storage fields and black fields are walls.

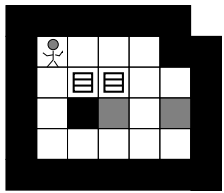

Fig. 4. A Sokoban Instance

We have written programs for DLV, GnT, Smodels and ASSAT (see Appendix A). These programs find solutions with a given number of push actions (where one push action can move a box over any number of fields) for a given puzzle, and we have developed scripts which iteratively run these programs with increasing numbers of push actions (starting at one) until some solution is found. In this way solutions with a minimal number of push actions are computed.

The puzzle in Figure 4 is solvable with six push actions, so each of the scripts uses the corresponding ASP system to first prove that no solutions with one to five push actions exist, and then to compute a solution with six push actions.

*Quantified Boolean Formulas (2QBF).* Our first benchmark problem at the second level of the Polynomial Hierarchy is 2QBF (see Section 3.3.5). The encoding for GnT is the same as for DLV, just using the different syntax for disjunction. This problem is $\Sigma_2^P$-complete, so we could not consider ASSAT and Smodels, since they do not support disjunction, which is strictly needed in this case (unless P = NP).

Our benchmark instances were generated following recent works presented in the literature that describe transition phase results for QBFs [Cadoli et al. 1997b; Gent and Walsh 1999]. They showed experimentally that classes of 2QBF instances with a certain #clauses/#variables ratio contain approximately the same number of valid and invalid instances, and are quite hard to be solved.

We generated randomly two data sets of 2QBF formulas $\Phi = \exists X \forall Y \phi$, where $\phi$ is in 3DNF. In all generated instances, the number of $\forall$-variables in any formula is the same as the number of $\exists$-variables (that is, $|X| = |Y|$) and each disjunct contains at least two universal variables. Moreover, in the first data set the number of clauses equals the overall number of variables (that is, $|X| + |Y|$), while in the second data set the number of clauses is $((|X| + |Y|)/2)^{0.5}$, as suggested by [Cadoli et al. 1997b] and [Gent and Walsh 1999], respectively. In the following, we will refer to instances generated according to the first schema as 2QBF$_{GW}$ and those generated according to the second schema as 2QBF$_{CGS}$.

Note that the above cited papers on phase transitions deal with dual 2QBF formulas of form $\Phi' = \forall X \exists Y \phi'$, where $\phi'$ is in 3CNF; for distinction, they are called 2QBF$_\forall^{\exists CNF}$ formulas and the former 2QBF$_\exists^{\exists DNF}$ formulas. Deciding the validity of 2QBF$_\forall^{\exists CNF}$ formulas is $\Pi_2^P$-complete. However, in our $\Sigma_2^P$-complete benchmark problem 2QBF, we have to decide the validity of 2QBF$_\exists^{\exists DNF}$ formulas. Nevertheless, our instance generation schemes are designed in such a way that the classes of QBFs from which we randomly select instances have the same phase transition behaviors as those described in [Cadoli et al. 1997b; Gent and Walsh 1999]. Indeed, it is easy to see that for each valid (resp., invalid) 2QBF$_\forall^{\exists CNF}$ formula $\Phi_c$, there is a corresponding invalid (resp., valid) 2QBF$_\exists^{\exists DNF}$ formula $\Phi_d$ with the same probability to be generated, and vice versa. Therefore, the fraction of valid instances (as a function of the #clauses/#variables ratio) for such 2QBF$_\forall^{\exists CNF}$ formulas is the same as the fraction of the invalid instances (as a function of the #disjuncts/#variables ratio) for our 2QBF$_\exists^{\exists DNF}$ formulas.

*Strategic Companies (STRATCOMP).* Here, we generated tests with instances for $n$ companies ($5 \leq n \leq 170$), $3n$ products, 10 uniform randomly chosen $contr\_by$ relations per company, and uniform randomly chosen $prod\_by$ relations.

To make the problem harder, we considered (up to) four producers per product, (up to) four controlling companies per company and only strategic sets containing two fixed companies (1 and 2, without loss of generality), slightly adjusting the program $\mathcal{P}_{strat}$ from Section 3.3.6 as follows:

$$strat(X1) \vee strat(X2) \vee strat(X3) \vee strat(X4) \coloneq$$
$$\quad prod\_by(X, X1, X2, X3, X4).$$
$$strat(W) \coloneq contr\_by(W, X1, X2, X3, X4),$$
$$\quad strat(X1), strat(X2), strat(X3), strat(X4).$$
$$\coloneq not\ strat(1).$$
$$\coloneq not\ strat(2).$$

We used the same encoding also for GnT (just rewriting the disjunction in GnT syntax). We could not consider ASSAT and Smodels, as they lack support for disjunction.

## 7.3 Results and Discussion

For HAMPATH, TSP, 2QBF$_{GW}$, 2QBF$_{CGS}$, and STRATCOMP, we generated 50 random instances for each problem size as indicated in the respective descriptions. For the more "deterministic" benchmarks REACH and SAMEGEN, we have one instance per size.

For every instance, we allowed a maximum running time of 7200 seconds (two hours) and a maximum memory usage of 256MB (using the command "limit datasize 256mega"); in the case of a system using an external instantiator, 256MB per individual component. In the graphs displaying the benchmark results, the line of a system stops whenever some problem instance was not solved within these time and memory limits.

In Figure 5, we show the results obtained for REACH (left) and SAMEGEN (right). Figures 6 and 7 show the results for 2QBF$_{GW}$ and 2QBF$_{CGS}$, respectively, while Figure 8 displays the results for STRATCOMP. The results for HAMPATH and TSP are shown in Figures 9 and 10, respectively. Tables IV and V, finally, contain the results for RAMSEY and SOKOBAN, respectively. In the graphs and the tables, for ASSAT, GnT, and Smodels, *native* denotes the usage of the Lparse instantiator, while DLV *Instantiator* denotes the runs of these systems when their input is grounded by DLV Instantiator (see Section 7.1).[15] For 2QBF$_{GW}$, 2QBF$_{CGS}$, STRATCOMP, and TSP, where we have 50 instances for each problem size, we report two graphs visualizing the average and the maximum computation time, respectively, consumed by each system over the 50 instances of the same size. For HAMPATH we provide two such pairs of graphs: the first for the "native" runs of the systems, the second for runs where DLV serves as the instantiator.

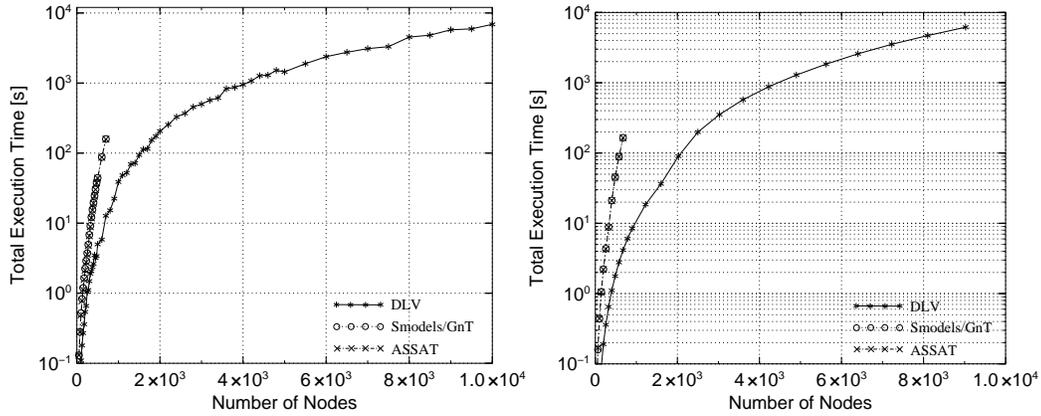

Fig. 5.   Reachability (REACH, left) and Same Generation (SAMEGEN, right)

On average, DLV outperforms the other systems on the set of proposed problems, but the results must be taken with some caveats, since they depend on the specific benchmark problems and, in some cases, also on the adopted encodings. We next discuss the results of the experiments, grouping similar cases together.

*Deductive Database Applications.*   On the deductive database applications REACH and SAMEGEN, DLV is significantly faster than the other systems, which is due to the in-stantiators used. Indeed these problems can be completely solved by the instantiators of all considered systems. Since Smodels, GnT, and ASSAT adopt the same instantiator (Lparse), they show essentially the same performance, differences are negligible and are due to minor architectural issues. All of them exhaust at 700 nodes on REACH and on 676 nodes on SAMEGEN, respectively, while DLV goes much further, up to 10000 nodes on REACH and up to 9025 nodes on SAMEGEN. Such a relevant difference in system performance is explained by the fact that Lparse does not employ database optimization techniques, and its instantiation method requires each variable in the program to be bound over a pre-defined domain. The instantiator of DLV, instead, incorporates several database techniques, and the domains are built dynamically, which often leads to a smaller result [Faber et al. 1999; Leone et al. 2001]; such a dynamic instantiation technique pays off in this case.

---

[15]The results for GnT with DLV Instantiator for 2QBF and STRATCOMP are not reported since there was basically no difference compared to Lparse. For the other benchmarks, DLV Instantiator was run with DLV encodings where for ASSAT and Smodels disjunction (only appearing in the "guessing rules") is translated by unstratified negation.

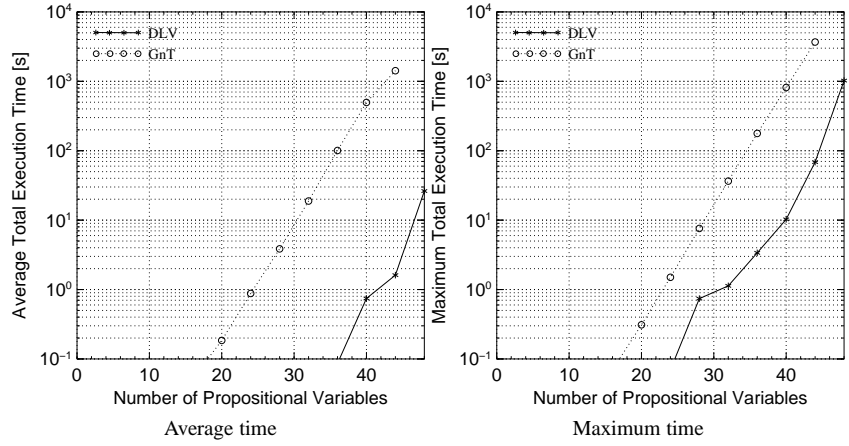

Fig. 6.    $2\text{QBF}_{GW}$

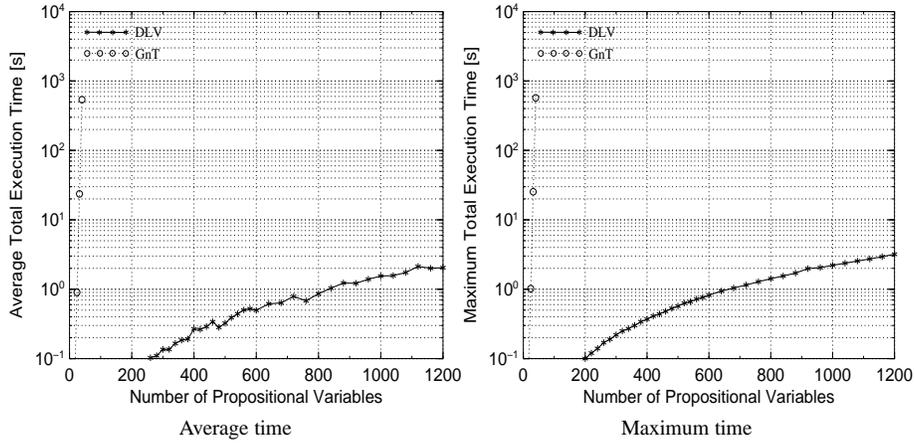

Fig. 7.    $2\text{QBF}_{CGS}$

$\Sigma_2^P$ *Problems*[16]. On $2\text{QBF}_{GW}$, both DLV and GnT scale in a very similar way, with DLV having a clear lead. In the case of $2\text{QBF}_{CGS}$ the difference between the two systems becomes significantly larger: the GnT system stops at size 40, while DLV easily solves all instances up to size 1200 (where we stopped benchmarking).

On the smaller instances of STRATCOMP, DLV and GnT behave similarly, but DLV scales better and reaches 170 companies, while GnT stops at 160 companies.

The higher efficiency of DLV on $\Sigma_2^P$-complete problems is not surprising, as the treatment of disjunction in DLV is built-in, while GnT employs a rewriting technique.

---

[16]Recall that Smodels and ASSAT are not considered here, since they do not support disjunction, which is strictly needed to solve $\Sigma_2^P$-complete problems like $2\text{QBF}_{GW}$ and STRATCOMP.

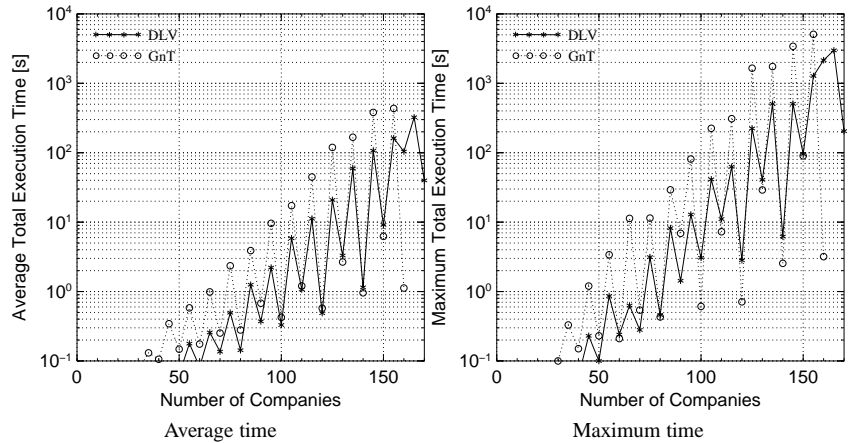

Average time                Maximum time

Fig. 8. Strategic Companies (STRATCOMP)

*HAMPATH and TSP.* On HAMPATH, the picture for DLV and GnT is similar to the one for $2QBF_{GW}$, which shows better performance of DLV. The results for the other systems are more surprising, however: DLV solves all instances up to size 105, while Smodels stops at size 50, and GnT and ASSAT stop at size 45.

Both Smodels and ASSAT are specialized for problems in NP, and we are aware that both are rather efficient on several NP-complete problems. In the literature, benchmark results on Hamiltonian Path or Hamiltonian Circuit have been reported also in [Lin and Zhao 2002] and [Nicolas et al. 2002]. While the findings in [Nicolas et al. 2002] appear to confirm our results, those in [Lin and Zhao 2002] are quite different. Apparently, this difference is mainly due to the encodings employed in the benchmarks. Indeed, we have tested the Hamiltonian Path programs of our benchmarks on the instances used in [Lin and Zhao 2002], and we arrived at similar results as reported above.

Interestingly, when we use DLV as instantiator, Smodels and GnT stop earlier (and solve all instances up to size 40 versus size 45 with Lparse), while ASSAT reaches size 45. This is somewhat unexpected, since the instantiation produced by the DLV instantiator is, in general, a subset of that generated by Lparse (on these examples they often coincide). The difference is probably due to the different orderings of the ground rules generated by the two instantiators, which may have an impact on the heuristics of GnT and Smodels in this case.

Finally, we observe that Smodels behaves better than ASSAT on small instances, and that the two systems scale similarly.

The picture for TSP is different:[17] the optimality of a solution makes the problem harder than HAMPATH. Both DLV and Smodels get significantly slower and have very similar run time, with DLV being slightly faster. This behavior suggests that the implementation of the optimization constructs is still to be improved in both DLV and Smodels.

*RAMSEY.* This problem highlights some interesting aspects related to the instantiation. First of all, the choice of the domain predicate (*node* vs *arc*) has a tremendous impact on

---

[17] Recall that ASSAT and GnT can not be tested on this problem, since they have no optimization constructs.

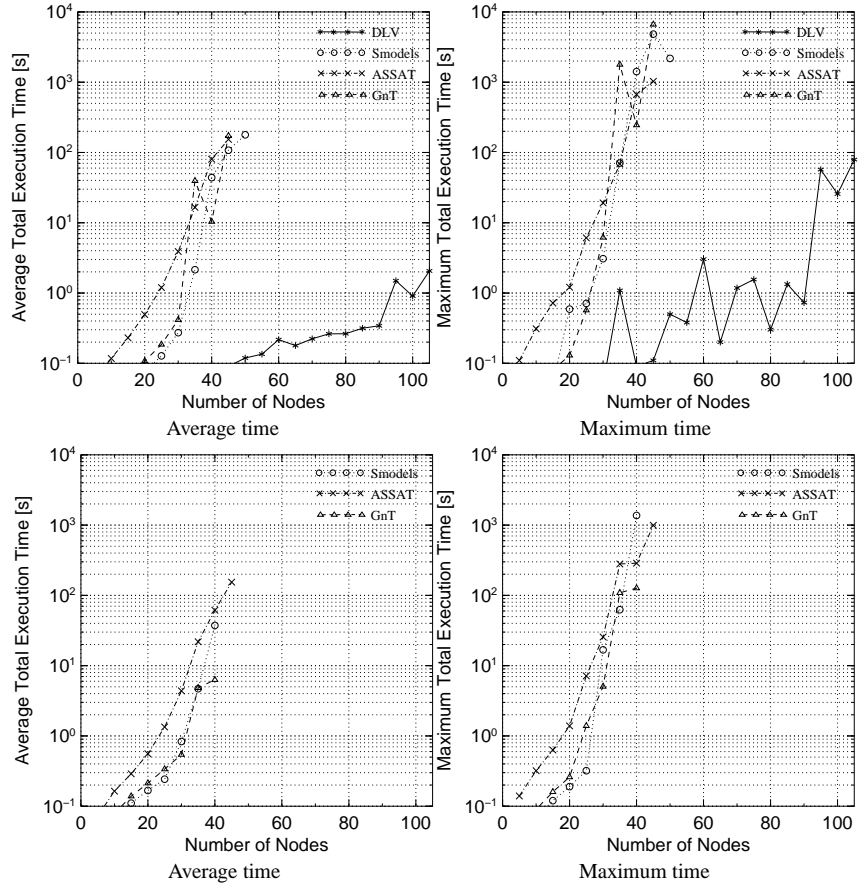

Fig. 9.   Hamiltonian Path (HAMPATH) – "native" (top), DLV Instantiator (bottom)

the performance of Lparse.[18]  Indeed, the three systems based on Lparse (GnT, Smodels, and ASSAT) behave significantly worse than DLV when $node$ is used as domain predicate, which is due to the ill-conditioned performance of Lparse on these encodings. Lparse generates too many ground instances on this program and requires a huge amount of memory. For instance, to prove that RAMSEY3-5 is not 13, the Intelligent Grounding module of DLV consumes 6MB of memory and generates 1,651 ground rules (including constraints); on the same instance, Lparse requires 42.5MB of memory and generates 373,737 ground rules.  Thus, the instantiation by Lparse is two orders of magnitude larger than the one by DLV, resulting in a huge overhead for GnT, Smodels and ASSAT. On larger instances (including all of RAMSEY3-6, RAMSEY3-7, and RAMSEY4-5), Lparse could not make the instantiation in the allowed amount of memory (256MB) and was halted. On the other hand, for the encoding with $edge$ as domain predicate, Lparse generates an instantiation of RAMSEY3-5 (13) with 1,820 ground rules using only 1.1MB of memory. Indeed, the per-

_______________

[18]The instantiator of DLV does not need the addition of domain predicates and, if domain predicates are added, their impact is very marginal on the instantiation time and they do not affect the size of the generated instantiation.

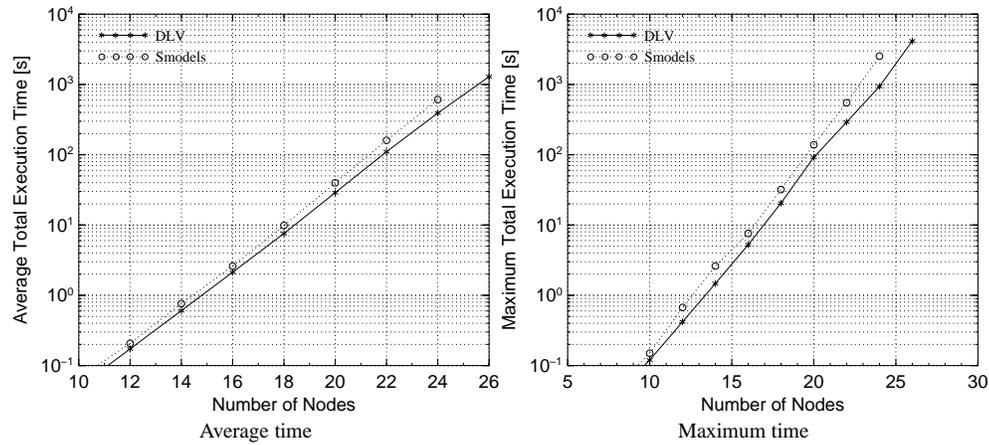

Fig. 10. Traveling Salesperson (TSP)

| Problem | DLV | GnT | | | Smodels | | | ASSAT | | |
|---|---|---|---|---|---|---|---|---|---|---|
| Size | | (node) | (edge) | (DLV) | (node) | (edge) | (DLV) | (node) | (edge) | (DLV) |
| 3-3 / 5 | 0.01 | 0.02 | 0.01 | 0.02 | 0.01 | 0.01 | 0.03 | 0.05 | 0.04 | 0.05 |
| 3-3 / 6 | 0.00 | 0.03 | 0.02 | 0.03 | 0.02 | 0.01 | 0.02 | 0.04 | 0.04 | 0.05 |
| 3-4 / 8 | 0.01 | 0.28 | 0.02 | 0.06 | 0.24 | 0.01 | 0.05 | 0.28 | 0.04 | 0.07 |
| 3-4 / 9 | 0.78 | 1.26 | 0.81 | 0.84 | 0.95 | 0.52 | 0.55 | 2.09 | 1.64 | 1.53 |
| 3-5 / 12 | 0.18 | 24.06 | 0.19 | 0.51 | 20.79 | 0.16 | 1.12 | 21.84 | 0.18 | 1.15 |
| 3-5 / 13 | 0.30 | 36.18 | 0.30 | 0.82 | 31.24 | 0.26 | 1.89 | 32.92 | 0.25 | 1.92 |
| 3-5 / 14 | - | - | - | - | - | - | - | 38.06 | 2032.30 | 1874.78 |
| 3-6 / 16 | 3.56 | - | 6.40 | 8.94 | - | 5.60 | 44.89 | - | 4.76 | 43.38 |
| 3-6 / 17 | 6.66 | - | 11.45 | 14.91 | - | 10.45 | 73.59 | - | 7.28 | 70.75 |
| 3-6 / 18 | - | - | - | - | - | - | - | - | 435.56 | 592.42 |
| 3-7 / 15 | 4.18 | - | 49.41 | 9.22 | - | 48.66 | 87.59 | - | 47.96 | 86.84 |
| 3-7 / 16 | 7.91 | - | 85.52 | 17.68 | - | 82.65 | 157.05 | - | 81.30 | 157.49 |
| 3-7 / 17 | 14.33 | - | 144.88 | - | - | 138.11 | - | - | 134.26 | - |
| 3-7 / 18 | 25.41 | - | 239.83 | - | - | - | - | - | 215.64 | - |
| 3-7 / 19 | 43.84 | - | 387.14 | - | - | - | - | - | 337.20 | - |
| 3-7 / 20 | 73.22 | - | 608.51 | - | - | 563.44 | - | - | 513.62 | - |
| 3-7 / 21 | 118.12 | - | 909.07 | - | - | 848.26 | - | - | 767.27 | - |
| 3-7 / 22 | 1476.08 | - | 1668.76 | - | - | - | - | - | 1125.67 | - |
| 3-7 / 23 | - | - | - | - | - | - | - | - | 2017.22 | - |
| 4-4 / 13 | 0.19 | 3.68 | 0.16 | 0.54 | 69.78 | 0.16 | 0.54 | 3.35 | 0.15 | 0.53 |
| 4-4 / 14 | 0.29 | 4.99 | 0.23 | 0.77 | 109.83 | 6.03 | 0.77 | 4.51 | 0.18 | 0.77 |
| 4-4 / 15 | 0.45 | 6.84 | 0.36 | 1.14 | - | 1.64 | 1.10 | 5.94 | 0.24 | 1.07 |
| 4-4 / 16 | 0.81 | 48.14 | 131.19 | 8.75 | - | 5.01 | - | 10.38 | 0.90 | 3.88 |
| 4-4 / 17 | 4282.26 | 98.32 | 1.45 | 28.19 | - | - | - | 13.80 | 1.07 | 4.57 |
| 4-4 / 18 | - | - | - | - | - | - | - | 723.88 | 492.69 | 588.68 |
| 4-5 / 17 | 2.50 | - | 2.66 | 6.02 | - | 1.83 | 12.70 | - | 1.27 | 11.99 |
| 4-5 / 18 | 3.78 | - | 4.12 | 8.80 | - | 11.09 | 18.82 | - | 1.73 | 18.09 |
| 4-5 / 19 | 5.55 | - | 6.11 | 12.64 | - | 988.37 | 27.13 | - | 2.34 | 26.24 |
| 4-5 / 20 | 7.56 | - | 7.99 | 17.56 | - | - | 39.64 | - | 3.16 | 35.99 |
| 4-5 / 21 | 10.74 | - | 11.71 | 28.14 | - | - | 55.86 | - | 4.13 | 51.11 |
| 4-5 / 22 | 14.91 | - | 39.91 | 34.81 | - | - | 83.22 | - | 5.57 | 68.53 |
| 4-5 / 23 | 53.93 | - | 96.31 | - | - | - | - | - | 12.56 | - |
| 4-5 / 24 | - | - | 182.81 | - | - | - | - | - | 182.81 | - |
| % solved | 84.84 | 33.33 | 84.84 | 63.63 | 24.24 | 60.60 | 57.57 | 39.39 | 100.00 | 72.72 |

Table IV. Results for RAMSEY (times in seconds)

| Problem# | native | | | | DLV Instantiator | | |
|---|---|---|---|---|---|---|---|
| | DLV | GnT | Smodels | ASSAT | GnT | Smodels | ASSAT |
| 1 | 15.41 | - | - | - | 75.13 | 23.78 | 6.36 |
| 2 | 0.16 | - | - | - | 0.71 | 0.61 | 0.80 |
| 3 | 0.21 | - | - | - | 1.12 | 0.88 | 0.97 |
| 4 | 0.13 | 55.71 | 49.05 | 58.36 | 0.69 | 0.58 | 0.75 |
| 5 | 0.32 | - | - | - | 1.53 | 1.25 | 1.73 |
| 6 | 0.11 | 34.49 | 30.43 | 32.40 | 0.48 | 0.42 | 0.53 |
| 7 | 0.15 | 84.43 | 74.44 | 70.66 | 0.77 | 0.62 | 0.75 |
| 8 | 0.46 | - | - | - | 2.13 | 1.74 | 2.00 |
| 9 | 0.09 | - | 99.32 | 94.25 | 0.43 | 0.36 | 0.43 |
| 10 | 0.18 | - | - | - | 0.84 | 0.69 | 0.79 |
| 11 | 0.15 | - | 155.04 | 145.69 | 0.69 | 0.58 | 0.69 |
| 12 | 0.17 | - | 156.24 | 186.24 | 0.80 | 0.68 | 0.80 |
| 13 | 0.06 | 13.20 | 11.52 | 11.34 | 0.33 | 0.28 | 0.35 |
| 14 | 0.11 | 34.54 | 30.47 | 32.04 | 0.81 | 0.68 | 0.84 |
| 15 | 0.17 | - | - | - | 0.58 | 0.48 | 0.57 |
| 16 | 1.14 | - | - | - | 5.69 | 4.09 | 4.41 |
| 17 | 0.14 | - | 109.28 | 119.88 | 0.71 | 0.58 | 0.71 |
| 18 | 0.92 | - | - | - | 5.72 | 3.68 | 3.08 |
| 19 | 0.59 | - | - | - | 5.20 | 3.22 | 2.87 |
| 20 | 0.76 | - | - | - | 4.48 | 3.19 | 2.73 |
| 21 | 0.27 | - | 157.43 | 162.44 | 1.23 | 1.01 | 1.16 |
| 22 | 0.27 | - | 223.57 | 209.98 | 1.18 | 0.98 | 1.11 |
| 23 | 0.29 | - | 224.04 | - | 1.34 | 1.07 | 1.45 |
| 24 | 0.26 | - | 157.67 | 170.00 | 1.24 | 1.00 | 1.33 |
| 25 | 0.08 | 11.35 | 9.91 | 10.72 | 0.43 | 0.36 | 0.47 |
| 26 | 0.38 | - | 213.14 | 248.01 | 1.94 | 1.56 | 1.93 |
| 27 | 0.06 | 13.26 | 11.58 | 11.23 | 0.32 | 0.27 | 0.36 |
| 28 | 0.34 | - | - | - | 1.51 | 1.20 | 1.55 |
| 29 | 0.54 | - | 191.14 | 241.92 | 3.14 | 2.28 | 2.66 |
| 30 | 0.39 | - | - | - | 1.66 | 1.32 | 1.52 |
| 31 | 0.53 | - | - | - | 2.74 | 1.89 | 2.12 |
| 32 | 0.06 | 13.41 | 11.69 | 12.74 | 0.35 | 0.29 | 0.37 |
| 33 | 0.31 | - | - | - | 1.46 | 1.20 | 1.46 |
| 34 | 0.49 | - | - | - | 1.96 | 1.59 | 1.99 |
| 35 | 1.00 | - | - | - | 4.71 | 3.27 | 4.06 |
| 36 | 0.20 | 79.70 | 70.27 | 77.99 | 1.02 | 0.83 | 1.08 |
| 37 | 0.66 | - | - | - | 3.07 | 2.33 | 2.26 |
| 38 | 0.46 | - | 215.15 | 271.75 | 2.04 | 1.61 | 2.08 |
| 39 | 0.43 | - | - | - | 1.87 | 1.49 | 2.03 |
| 40 | 0.27 | - | 157.08 | 181.74 | 1.19 | 0.99 | 1.37 |
| 41 | 0.32 | - | 145.70 | 153.56 | 1.56 | 1.26 | 1.55 |
| 42 | 0.31 | 107.89 | 95.94 | 109.38 | 1.48 | 1.22 | 1.52 |
| 43 | 0.74 | - | - | - | 4.07 | 2.88 | 2.82 |
| 44 | 0.38 | - | 214.17 | 224.76 | 1.64 | 1.34 | 1.58 |
| 45 | 0.49 | - | 191.05 | 218.84 | 2.32 | 1.88 | 2.19 |
| 46 | 0.69 | - | 353.98 | - | 4.05 | 2.78 | 3.77 |
| 47 | 0.96 | - | - | - | 5.36 | 3.32 | 3.24 |
| 48 | 0.89 | - | 354.23 | - | 4.17 | 3.06 | 3.38 |
| 49 | 0.41 | - | - | - | 1.85 | 1.52 | 1.63 |
| 50 | 659.02 | - | - | - | 233.07 | 370.21 | 12.00 |
| 51 | 198.84 | - | - | - | - | 2033.88 | 263.88 |
| 52 | - | - | - | - | - | 5631.03 | 234.42 |
| 53 | 0.56 | - | 279.91 | 322.20 | 3.32 | 2.40 | 2.71 |
| 54 | 12.31 | - | - | - | 163.51 | 45.96 | 13.00 |
| 55 | 320.44 | - | - | - | 196.89 | 122.37 | 129.66 |
| 56 | 2.70 | - | - | - | 13.08 | 6.86 | 6.65 |
| 57 | - | - | - | - | - | - | - |
| 58 | - | - | - | - | - | - | - |
| 59 | 19.98 | - | - | - | 106.95 | 104.58 | 19.55 |
| 60 | 5.27 | - | - | - | 173.32 | 25.65 | 19.32 |
| % solved | 95.00 | 16.67 | 46.67 | 41.67 | 93.33 | 96.67 | 96.67 |

Table V. Results for SOKOBAN (times in seconds)

formance of GnT, Smodels and ASSAT changes completely on the encodings with $edge$, and the three systems perform significantly better than on the encoding with $node$.[19] The performance of GnT comes closer to the performance of DLV, and ASSAT outperforms all other systems on increasing instances.

Surprisingly, coupling ASSAT, GnT, and Smodels with the DLV instantiator does not work well in this case, revealing an inefficiency of DLV in the instantiation of this specific non-disjunctive encoding (the resulting instantiation is small, but instantiation time is long).

*SOKOBAN.* Also on this benchmark, the instantiation process has a strong impact on overall system performance. If the native instantiators are used, DLV significantly outperforms the other systems; the difference is mainly due to the instantiation modules. Lparse requires much more memory and generates a higher number of rule instances than DLV. For instance, on SOKOBAN #48 (with plan length 8), DLV requires 6MB of memory and generates 3,236 ground rules, while Lparse requires 125MB of memory and generates 2,130,705 ground rules. In fact, in many cases Lparse exceeded the memory limit and was halted.

We see a very different picture when the DLV instantiator is used. ASSAT, GnT and Smodels perform like DLV; in fact, ASSAT and Smodels solve even one instance more than DLV now. The encoding of SOKOBAN is much longer and more complex than the other benchmark programs, some optimization techniques employed by the DLV instantiator improve the "quality" (i.e., the size) of the instantiation with a great benefit for the performance of the other systems.

*Summary.* Summarizing, we can draw the following conclusions:

—Currently, the DLV system is better suited to deal with "database oriented" applications (in the sense of deductive databases) than the other systems, as the optimization techniques implemented in its instantiator allow DLV to handle larger amounts of data in a more efficient way.

—The built-in implementation of disjunction allows the DLV system to be faster than the other systems on hard ($\Sigma_2^P$-hard) problems. Even if GnT is slower on such problems, its performance is very good considering that disjunction is implemented through a rewriting technique. Smodels and ASSAT cannot handle $\Sigma_2^P$-hard problems at all, since they do not support disjunction and $\Sigma_2^P$-hard problems cannot be encoded uniformly by v-free programs, unless the Polynomial Hierarchy collapses (cf. Section 4).

—On NP/co-NP/$\Delta_2^P$ problems like HAMPATH and TSP, the performance of the systems comes closer, and they seem to be more strongly influenced by the chosen encodings and by the instantiation procedures. DLV has a lead on problems involving recursion like HAMPATH (with the chosen encoding being relevant here), while ASSAT is best on RAMSEY, where the search space is very large and the computationally expensive heuristics of DLV and Smodels (based on look-ahead) appear unsuited. On SOKOBAN DLV outperforms the other systems if the native instantiators are used, while the other systems improve significantly when they adopt the DLV instantiator.

---

[19]We cannot explain the performance of GnT around RAMSEY4-4/16, but we verified that the timings are indeed correct.

—Thanks to the good performance on deductive database applications and on $\Sigma_2^P$-complete problems, DLV appears to be applicable to a larger range of problems than the other systems, which have been designed for a more specific class of problems.

—The approach implemented in the recent ASSAT system is very interesting, as it allows to exploit the huge amount of work on satisfiability checkers done in AI. This approach is promising especially on NP/co-NP problems, where the well-assessed heuristics of satisfiability checkers pay off (e.g., ASSAT was fast on the RAMSEY problem). However, this system still needs some work to make it more robust; we experienced a number of crashes on the Sokoban encodings, which are more elaborated than the others.
It would be nice to see an extension of ASSAT for dealing with optimization problems, and enhancing ASSAT to compute *all* answer sets seems very important. Indeed, while DLV, GnT, and Smodels are able to compute all (or a given number of) answer sets of a program, ASSAT cannot generate all answer sets in one computation (see Section 7.1). This limitation may have a negative impact on the efficiency of ASSAT on some relevant tasks like, for instance, computing the set of all brave or cautious consequences.

—Comparing the use of disjunction versus unstratified negation or other means to express choice, the experiments showed that, in the specific benchmark problems considered, DLV, and in some cases also GnT, is competitive to Smodels and ASSAT, and the use of the more familiar disjunction connective does not come at higher computational price.

Two strong points of DLV in comparison to the other systems are: (i) the built-in implementation of disjunction, which allows DLV to efficiently solve $\Sigma_2^P$-complete problems, and (ii) its advanced instantiation module, which allows DLV to efficiently deal with deductive database applications, which supports a simpler way of programming (the user does not have to bother about domain predicates), and which also tends to affect performance on hard problems (like SOKOBAN) positively, as it produces a smaller instantiation than Lparse, and requires less memory.

Our experimental results are complemented by other efforts on benchmarking ASP systems recently reported in the literature [Dix et al. 2002; Koch and Leone 1999; Lin and Zhao 2002; Nicolas et al. 2002; Simons et al. 2002; Eiter et al. 1998a; Janhunen et al. 2003]. In general, these confirm our claim that DLV handles deductive database applications and $\Sigma_2^P$-complete problems more efficiently than the other systems, while the situation is less clear on NP-complete problems where the results depend, to a larger extent, on the encodings adopted. While the experiments in [Koch and Leone 1999; Lin and Zhao 2002; Nicolas et al. 2002; Simons et al. 2002; Eiter et al. 1998a] are conducted on benchmark problems very similar to those used in this paper, a completely different approach has been taken in [Dix et al. 2002]. This paper reports on a most recent comprehensive benchmarking activity which was independently performed in the context of Hierarchical Task Network (HTN) Planning based on ASP. Dix *et al.* considered instances of several planning problems, which are transformed to logic programs and solved by invoking an answer set engine. These experiments are relevant, as they compare the systems in the important domain of AI planning on benchmark instances which are really used to compare planning systems. In particular, Dix *et al.* used DLV and Smodels (version v2.27) and compared the behaviors of the systems. Summarizing their findings in [Dix et al. 2002], the authors state that DLV was significantly faster than Smodels in their experiments. They believe one of the reasons for this is grounding, as Smodels requires domain predicates, creating many ground instances of the program clauses which are often irrelevant. But Dix *et al.*

also conclude that the better grounding is not the only source for DLV's superiority in their experiments: DLV was still faster than Smodels running on an instantiation provided by DLV.

## 8.  CONCLUSION

After an extensive period of mainly foundational and theoretical research on nonmonotonic logics and databases, during the last years several implementations became available which can be utilized as advanced and powerful tools for problem solving in a highly declarative manner. They provide a computational back-end for the Answer Set Programming (ASP) paradigm, in which the solutions to a problem are encoded in the models (or answer sets) of a nonmonotonic theory.

In this paper, we have presented the DLV system, which is a state-of-the-art implementation of disjunctive logic programming under the answer set semantics [Gelfond and Lifschitz 1991], enriched by further useful language constructs. This paper is the first comprehensive document in which a wide survey over several relevant aspects of the system is provided, covering technical, methodological, and application aspects.

Starting from an exposition of the core system language, we have illustrated how knowledge representation and reasoning problems, even of high computational complexity, can be declaratively encoded in the DLV language following a Guess-Check-Optimize (**GCO**) paradigm. Furthermore, we have addressed computational aspects of the language by providing a complete picture of the complexity of the DLV language and relevant syntactic fragments thereof. The DLV engine exploits this picture by handling fragments of lower complexity more efficiently. On the benchmarking side, as an important contribution this paper presents and discusses a thorough experimentation activity which we have carried out on an ample set of benchmark problems, taken from various application areas and with different computational complexity. These experiments provide some evidence that, in comparison with similar systems, DLV may have a wider range of applicability and, due to its built-in database techniques and sophisticated processing of non-ground programs, is better suited to deal with larger amounts of input data. Furthermore, by its high expressiveness (up to $\Delta_3^P$-complete problems), it is capable of representing more complex problems than comparable systems, whose expressiveness is limited to lower classes in the Polynomial Hierarchy.

The DLV system is disseminated in academia, and presumably soon also in industry. Indeed, the industrial exploitation of DLV in the emerging areas of Knowledge Management and Information Integration is the subject of two international projects funded by the European Commission, namely, INFOMIX (Boosting Information Integration, project IST-2002-33570) and ICONS (Intelligent Content Management System, project IST-2001-32429). Further research on the DLV system is pursued in a number of national projects such as the FWF (Austrian Science Funds) project P14781 on logic-based planning and in the Ludwig Wittgenstein Laboratory of Information Systems (FWF project Z29-N04).

DLV is widely used for educational purposes in courses on databases and on AI, both in European and American universities. The DLV system has been employed at CERN, the European Laboratory for Particle Physics located near Geneva, for an advanced deductive database application that involves complex knowledge manipulation on large-sized databases. The Polish company Rodan Systems S.A. exploits DLV in a tool for the detection of price manipulations and unauthorized use of confidential information, which is used by the Polish Securities and Exchange Commission. We believe that the strengths of DLV – its expressivity and solid implementation – make it attractive for similar applications.

The DLV system has been continuously extended and improved over the last years, and several language enhancements are under development which will be available in future releases. In particular, aggregates and other set functions will be available [Dell'Armi et al. 2003], which are very convenient for the encoding of many practical problems. On the other hand, improvements at the algorithmic level are underway by the development and implementation of more sophisticated magic set techniques than those which are currently available for disjunctive logic programs [Greco 1999]. Furthermore, refinements of the heuristics in the model generation process should improve performance and scalability.

From what has been achieved and what may be expected, we are confident that DLV, as well as similar ASP systems, will provide us with more and more widely applicable tools which render computational logic an important component in advanced information technology. ASP has been recognized as a promising approach for dealing with problems which require advanced modeling capabilities for problem representation – recently, the European Commission granted funding for a special Working Group on Answer Set Programming (WASP) that is formed by a number of research groups in Europe.

ACKNOWLEDGMENTS

We would like to thank Robert Bihlmeyer, Francesco Buccafurri, Francesco Calimeri, Simona Citrigno, Tina Dell'Armi, Giuseppe Ielpa, Christoph Koch, Cristinel Mateis, and Axel Polleres who contributed to the DLV project, as well as Ilkka Niemelä and Patrik Simons for fruitful discussions. Furthermore, we are grateful to Chitta Baral, Leo Bertossi, Jürgen Dix, Esra Erdem, Michael Fink, Michael Gelfond, Giuliana Sabbatini, Terry Swift, and many others for sharing their experiences on DLV with us and for useful suggestions. And finally, we would like to thank the anonymous reviewers for their valuable comments.

## A. APPENDIX: SOKOBAN ENCODINGS

The SOKOBAN problem has been encoded in DLV as follows.

$time(T) \text{ :- } \#int(T).$
$actiontime(T) \text{ :- } \#int(T), T <> \#maxint.$

$left(L1, L2) \text{ :- } right(L2, L1).$
$bottom(L1, L2) \text{ :- } top(L2, L1).$

$adj(L1, L2) \text{ :- } right(L1, L2).$
$adj(L1, L2) \text{ :- } left(L1, L2).$
$adj(L1, L2) \text{ :- } top(L1, L2).$
$adj(L1, L2) \text{ :- } bottom(L1, L2).$

$push(B, right, B1, T) \text{ v } \neg push(B, right, B1, T) \text{ :- } reachable(L, T), right(L, B), box(B, T),$
$\qquad pushable\_right(B, B1, T), good\_pushlocation(B1), actiontime(T).$
$push(B, left, B1, T) \text{ v } \neg push(B, left, B1, T) \text{ :- } reachable(L, T), left(L, B), box(B, T),$
$\qquad pushable\_left(B, B1, T), good\_pushlocation(B1), actiontime(T).$
$push(B, up, B1, T) \text{ v } \neg push(B, up, B1, T) \text{ :- } reachable(L, T), top(L, B), box(B, T),$
$\qquad pushable\_top(B, B1, T), good\_pushlocation(B1), actiontime(T).$
$push(B, down, B1, T) \text{ v } \neg push(B, down, B1, T) \text{ :- } reachable(L, T), bottom(L, B), box(B, T),$
$\qquad pushable\_bottom(B, B1, T), good\_pushlocation(B1), actiontime(T).$

$reachable(L, T) \text{ :- } sokoban(L, T).$
$reachable(L, T) \text{ :- } reachable(L1, T), adj(L1, L), \text{ not } box(L, T).$

$pushable\_right(B, D, T) \text{ :- } box(B, T), right(B, D), \text{ not } box(D, T), actiontime(T).$
$pushable\_right(B, D, T) \text{ :- } pushable\_right(B, D1, T), right(D1, D), \text{ not } box(D, T).$
$pushable\_left(B, D, T) \text{ :- } box(B, T), left(B, D), \text{ not } box(D, T), actiontime(T).$
$pushable\_left(B, D, T) \text{ :- } pushable\_left(B, D1, T), left(D1, D), \text{ not } box(D, T).$
$pushable\_top(B, D, T) \text{ :- } box(B, T), top(B, D), \text{ not } box(D, T), actiontime(T).$
$pushable\_top(B, D, T) \text{ :- } pushable\_top(B, D1, T), top(D1, D), \text{ not } box(D, T).$
$pushable\_bottom(B, D, T) \text{ :- } box(B, T), bottom(B, D), \text{ not } box(D, T), actiontime(T).$
$pushable\_bottom(B, D, T) \text{ :- } pushable\_bottom(B, D1, T), bottom(D1, D), \text{ not } box(D, T).$

$sokoban(L, T1) \text{ :- } push(\_, right, B1, T), \#succ(T, T1), right(L, B1).$
$sokoban(L, T1) \text{ :- } push(\_, left, B1, T), \#succ(T, T1), left(L, B1).$
$sokoban(L, T1) \text{ :- } push(\_, up, B1, T), \#succ(T, T1), top(L, B1).$
$sokoban(L, T1) \text{ :- } push(\_, down, B1, T), \#succ(T, T1), bottom(L, B1).$

$box(B, T1) \text{ :- } push(\_, \_, B, T), \#succ(T, T1).$
$\neg box(B, T1) \text{ :- } push(B, \_, \_, T), \#succ(T, T1).$

$box(LB, T1) \text{ :- } box(LB, T), \#succ(T, T1), \text{ not } \neg box(LB, T1).$

$\text{:- } push(B, \_, \_, T), push(B1, \_, \_, T), B <> B1.$
$\text{:- } push(B, D, \_, T), push(B, D1, \_, T), D <> D1.$
$\text{:- } push(B, D, B1, T), push(B, D, B11, T), B1 <> B11.$

$good\_pushlocation(L) \text{ :- } right(L, \_), left(L, \_).$
$good\_pushlocation(L) \text{ :- } top(L, \_), bottom(L, \_).$
$good\_pushlocation(L) \text{ :- } solution(L).$

$goal\_unreached \text{ :- } solution(L), \text{ not } box(L, \#maxint). \quad \text{:- } goal\_unreached.$

For GnT, we used the same encoding with the addition of domain predicates and minor syntactic changes to conform to the different syntax.

The encoding for Smodels and ASSAT has been obtained from the encoding for GnT, rewriting disjunction in the "guessing" rules for $push$ by unstratified negation.